\documentclass[11pt,a4paper]{article}

\usepackage{microtype}
\usepackage[english]{babel}

\usepackage[
  top=2.4cm,
  bottom=2.6cm,
  left=2.4cm,
  right=2.4cm,
]{geometry}

\usepackage{libertine}%
\usepackage{amsthm}%
\usepackage[libertine]{newtxmath}%
\usepackage[no-math]{fontspec}%
\newfontfamily\jbsans[
  Path           = fonts/,
  Extension      = .ttf,
  UprightFont    = JetBrainsSans-Regular,
  BoldFont       = JetBrainsSans-Bold,
  FontFace       = {l}{n}{JetBrainsSans-Light},
  FontFace       = {sb}{n}{JetBrainsSans-SemiBold},
  FontFace       = {eb}{n}{JetBrainsSans-ExtraBold},
]{JetBrainsSans}
\usepackage{fontawesome5}%
\usepackage{inconsolata}%
\usepackage[dvipsnames,svgnames,x11names]{xcolor}

\definecolor{mellum-primary}{HTML}{27282c}%
\definecolor{mellum-accent}{HTML}{fc801d}%
\definecolor{mellum-light}{HTML}{F0F4FA}%
\definecolor{mellum-mid}{HTML}{087cfa}%
\definecolor{mellum-gray}{HTML}{6B7B8D}%
\definecolor{mellum-dark}{HTML}{000000}%
\definecolor{mellum-code-bg}{HTML}{F7F8FB}%
\definecolor{mellum-link}{HTML}{ff318c}%
\definecolor{mellum-green}{HTML}{28b8a0}%
\definecolor{mellum-red}{HTML}{DC2626}%

\usepackage{graphicx}
\usepackage{float}
\usepackage{wrapfig}
\usepackage{subcaption}
\usepackage{tikz}
\usetikzlibrary{calc,positioning,arrows.meta,shapes.geometric,backgrounds,fit,patterns,decorations.pathreplacing}
\usepackage{pgfplots}
\pgfplotsset{compat=1.18}
\graphicspath{{./}{figures/}}

\usepackage{booktabs}
\usepackage{multirow}
\usepackage{colortbl}
\usepackage{tabularx}
\usepackage{array}
\usepackage{siunitx}
\newcommand{\tableheadercolor}{}

\usepackage{mathtools}
\usepackage{bm}

\usepackage[ruled,vlined,linesnumbered]{algorithm2e}
\SetAlgoLined
\SetKwComment{Comment}{$\triangleright$\ }{}

\usepackage{listings}
\lstdefinestyle{mellumcode}{
  backgroundcolor=\color{mellum-code-bg},
  basicstyle=\ttfamily\small,
  breakatwhitespace=false,
  breaklines=true,
  captionpos=b,
  commentstyle=\color{mellum-gray}\itshape,
  keywordstyle=\color{mellum-mid}\bfseries,
  stringstyle=\color{mellum-accent},
  numberstyle=\tiny\color{mellum-gray},
  numbers=left,
  numbersep=8pt,
  frame=none,
  xleftmargin=1.2em,
  framexleftmargin=1.2em,
  tabsize=2,
  showstringspaces=false,
  aboveskip=0.8em,
  belowskip=0.8em,
  rulecolor=\color{mellum-primary!20},
}
\definecolor{mellum-link-internal}{HTML}{6b57ff}%
\usepackage[
  colorlinks=true,
  linkcolor=mellum-link-internal,
  citecolor=mellum-link-internal,
  urlcolor=mellum-link,
  bookmarks=true,
  bookmarksnumbered=true,
]{hyperref}
\usepackage[capitalise,noabbrev]{cleveref}

\usepackage{csquotes}
\usepackage[
  backend=biber,
  style=numeric-comp,
  sorting=nyt,
  maxbibnames=99,
  giveninits=true,
  doi=false,
  isbn=false,
  url=true,
]{biblatex}
\usepackage{titlesec}

\titleformat{\section}
  {\Large\bfseries\color{mellum-primary}}
  {\color{mellum-accent}\thesection}{0.8em}{}

\titleformat{\subsection}
  {\large\bfseries\color{mellum-dark}}
  {\color{mellum-accent}\thesubsection}{0.6em}{}

\titleformat{\subsubsection}[hang]
  {\normalsize\bfseries\color{mellum-dark}}
  {\color{mellum-accent}\thesubsubsection}{0.6em}{}

\titlespacing*{\section}{0pt}{2.0ex plus 0.6ex minus .2ex}{1.0ex plus .2ex}
\titlespacing*{\subsection}{0pt}{1.4ex plus 0.4ex minus .2ex}{0.6ex plus .1ex}
\titlespacing*{\subsubsection}{0pt}{1.0ex plus 0.3ex minus .1ex}{0.4ex plus .1ex}

\newcommand{\reportdate}{May 2026}
\newcommand{\reportversion}{v1.0}

\usepackage{fancyhdr}
\renewcommand{\headrulewidth}{0.4pt}
\renewcommand{\headrule}{\hbox to\headwidth{\color{mellum-primary!40}\leaders\hrule height \headrulewidth\hfill}}
\fancypagestyle{firstpage}{
  \fancyhf{}
  \renewcommand{\headrulewidth}{0pt}
}

\renewcommand{\footnoterule}{%
  \kern-3pt
  {\color{mellum-primary!25}\hrule width 0.35\columnwidth height 0.3pt}%
  \kern 2.6pt}

\usepackage[framemethod=tikz]{mdframed}

\newmdenv[
  backgroundcolor=mellum-light,
  linecolor=mellum-primary,
  linewidth=0.5pt,
  roundcorner=2pt,
  innerleftmargin=10pt,
  innerrightmargin=10pt,
  innertopmargin=8pt,
  innerbottommargin=8pt,
  frametitlefont=\bfseries\color{mellum-primary},
  frametitleaboveskip=6pt,
  frametitlebelowskip=4pt,
]{keybox}

\newmdenv[
  backgroundcolor=mellum-accent!5,
  linecolor=mellum-accent,
  linewidth=0.5pt,
  roundcorner=2pt,
  innerleftmargin=10pt,
  innerrightmargin=10pt,
  innertopmargin=8pt,
  innerbottommargin=8pt,
  frametitlefont=\bfseries\color{mellum-accent!90!black},
  frametitleaboveskip=6pt,
  frametitlebelowskip=4pt,
]{accentbox}

\newmdenv[
  topline=false,
  bottomline=false,
  rightline=false,
  leftline=true,
  linecolor=mellum-accent,
  linewidth=2pt,
  backgroundcolor=mellum-light,
  innerleftmargin=12pt,
  innerrightmargin=12pt,
  innertopmargin=8pt,
  innerbottommargin=8pt,
  skipabove=10pt,
  skipbelow=10pt,
  leftmargin=8pt,
  rightmargin=8pt,
]{rolloutquoteframe}
\newenvironment{rolloutquote}
  {\begin{rolloutquoteframe}\itshape\small}
  {\end{rolloutquoteframe}}

\usepackage[
  font={small},
  labelfont={bf,sc,color=mellum-primary},
  textfont={color=mellum-dark},
  format=plain,
  labelsep=quad,
]{caption}
\DeclareCaptionLabelSeparator{emdash}{\,\textemdash\,}
\usepackage{enumitem}
\setlist[itemize]{leftmargin=1.5em, itemsep=0.2em, parsep=0.1em}
\setlist[enumerate]{leftmargin=1.5em, itemsep=0.2em, parsep=0.1em}
\setlist[itemize,1]{label={\color{mellum-accent}\textbullet}}
\setlist[itemize,2]{label={\color{mellum-mid}\textendash}}

\usepackage{xspace}
\usepackage{pifont}
\usepackage[normalem]{ulem}

\newcommand{\modelname}{\textsc{Mellum~2}\xspace}
\newcommand{\prevmodelname}{\textsc{Mellum}\xspace}

\begin{document}

\thispagestyle{firstpage}

\vspace*{-2.0cm}

\noindent\includegraphics[height=0.75cm]{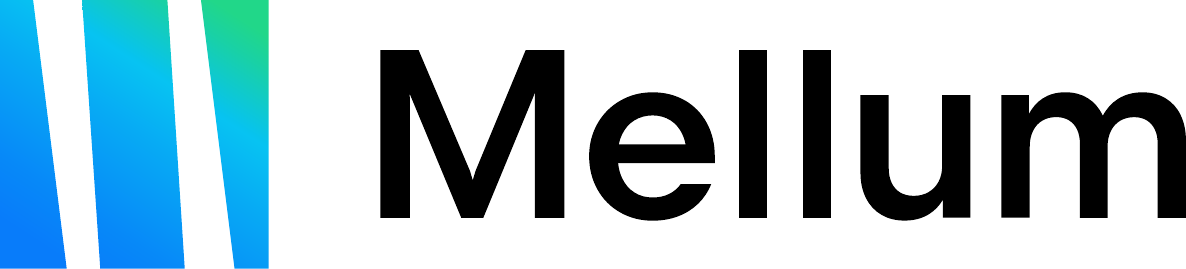}\par
\vspace{0.15cm}

\noindent\makebox[0pt][l]{\hspace*{-2.4cm}\textcolor{mellum-primary}{\rule{\paperwidth}{3.5pt}}}\par

\vspace{0.7cm}

\begin{center}

{
  \jbsans
  \fontsize{36}{42}\selectfont
  \color{mellum-primary}\textbf{Mellum 2}
}

\vspace{0.25cm}

{
  \jbsans
  \fontsize{14}{18}\selectfont
  \color{mellum-gray}Technical Report
}

\vspace{0.20cm}

{
  \small\color{mellum-accent}\textsf{\reportversion\,\textperiodcentered\,\reportdate}
}

\vspace{1.0cm}

{
  \normalsize\color{mellum-dark}

  Marko Kojic\textsuperscript{1}\quad
  Ivan Bondyrev\textsuperscript{1}\quad
  Aral de Moor\textsuperscript{1}\quad
  Joseph Shtok\textsuperscript{1}\\[0.35em]
  Petr Borovlev\textsuperscript{1,2}\quad
  Kseniia Lysaniuk\textsuperscript{1,2}\quad
  Madeeswaran Kannan\textsuperscript{1}\quad
  Ivan Dolgov\textsuperscript{1}\\[0.35em]
  Nikita Pavlichenko\textsuperscript{1}

  \vspace{0.5cm}

  {\small\color{mellum-gray}\textsuperscript{1}JetBrains\quad\textsuperscript{2}Constructor University, Bremen, Germany}
}

\vspace{0.9cm}

\begin{minipage}{0.92\textwidth}
  \setlength{\parindent}{0pt}

  \begin{keybox}[frametitle=Abstract]
    We present \modelname, an open-weight 12B-parameter Mixture-of-Experts
    (MoE) language model with 2.5B active parameters per token. \modelname is
    a general-purpose language model specialized in software engineering, spanning code generation and editing, debugging, multi-step reasoning, tool use and function calling, agentic coding, and conversational programming assistance, and it is the successor to the completion-focused 4B dense \prevmodelname model. The architecture builds on the
    Mixture-of-Experts (64 experts, 8 active) and combines
    Grouped-Query Attention with 4 KV heads, Sliding Window Attention on
    three of every four layers, and a single Multi-Token Prediction head
    that doubles as both an auxiliary pre-training objective and a built-in
    draft model for speculative decoding; each choice was validated by
    ablation with inference efficiency on commodity GPUs as a 
    design constraint. Pre-training spans approximately 10.6 trillion tokens
    through a three-phase curriculum that progressively shifts the mixture
    from diverse web data toward curated code and mathematical content
    (code ratio 23\,\%\,$\to$\,42\,\%\,$\to$\,59\,\%), optimized with Muon under
    FP8 hybrid precision and a Warmup-Hold-Decay schedule with linear
    decay to zero. The pre-trained base is extended to a 128K context window
    via a layer-selective YaRN and then post-trained in two stages
    (supervised fine-tuning followed by reinforcement learning with
    verifiable rewards), yielding two released variants: an \emph{Instruct}
    model that answers directly and a \emph{Thinking} model that emits an
    explicit reasoning trace before its final answer. Across code generation,
    math and reasoning, tool use, knowledge, and safety benchmarks,
    \modelname is competitive with open-weight baselines in the 4B--14B range
    while running at the per-token compute of a 2.5B dense model. We release
    the base, instruct, and thinking checkpoints, together with this report
    on the architecture decisions, data pipeline, and training recipe behind
    them, under the Apache~2.0 license.
  \end{keybox}

\end{minipage}

\vspace{0.5cm}

{
  \small\sffamily
  \href{https://huggingface.co/collections/JetBrains/mellum-2}{%
    \colorbox{mellum-accent!10}{\color{mellum-accent!90!black}\strut\,%
      \raisebox{-0.15em}{\includegraphics[height=1em]{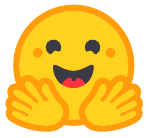}}\;Hugging Face\,}}%
  \hspace{0.4em}%
  \href{https://blog.jetbrains.com/ai/2026/05/mellum2-goes-open-source-a-fast-model-for-ai-workflows}{%
    \colorbox{mellum-primary!8}{\color{mellum-primary}\strut\,\faIcon{newspaper}\;Blog post\,}}%
  \hspace{0.4em}%
  \colorbox{mellum-light}{\color{mellum-primary}\strut\,\faIcon{balance-scale}\;Apache~2.0\,}%
}

\vspace{0.8cm}

\end{center}

\par\vspace*{\fill}
{\small\color{mellum-gray}\noindent
Correspondence: \texttt{mellum@jetbrains.com}\hfill Released under the Apache~2.0 license.}
\par\vspace{6pt}
\noindent\makebox[0pt][l]{\hspace*{-2.4cm}\textcolor{mellum-primary}{\rule{\paperwidth}{3.5pt}}}\par
\vspace*{-1ex}

\clearpage

\section{Introduction}
\label{sec:introduction}

Large language models (LLMs) have reshaped how developers work with code.
What began as inline autocomplete~\cite{li2023starcoder} has broadened into
a much wider task surface: writing whole functions from
natural-language specifications, editing and debugging existing code,
reasoning through multi-step engineering tasks, calling external tools,
navigating repositories as an agent, and serving as a conversational
collaborator throughout the development loop. The competitive coding models
today must do all of this at once, and at a serving cost that makes them
practical to deploy at scale.

Two regimes dominate the open-weights landscape on the quality-versus-cost
trade-off. Dense models in the 4--14B range are cheap to serve but plateau on
harder coding and reasoning workloads; very large Mixture-of-Experts (MoE)
models~\cite{fedus2022switch,dai2024deepseekmoe} reach frontier quality but at
deployment costs that strain everyday use. To strike a balance between knowledge scope and serving cost, we aim to extend the recent line of small
MoE coding models---among them Qwen3-Coder-30B-A3B~\cite{yang2025qwen3} and Ling-Coder-Lite~\cite{codefuse2025lingcoderlite}: sufficient parameters to absorb the long tail of programming language and reasoning knowledge but with enough sparsity to allow for deployment on commodity hardware (per-token compute in the 2--3B-dense range).

We introduce \modelname, an open-weight 12B-parameter MoE language model
with 2.5B active parameters per token, a general-purpose successor to
\prevmodelname~\cite{mellum2025} --- the 4B dense code-completion model previously
deployed in JetBrains IDEs. While the original \prevmodelname was trained to fill
single completions inside an editor, \modelname is a full-fledged coding
assistant: it generates and edits code, calls tools, plans and executes
multi-step agentic workflows, holds long conversations about code, and, in
its thinking variant, produces explicit reasoning traces before answering.
The model is built on the Qwen3-MoE recipe~\cite{yang2025qwen3} (64 experts,
8 active) with three deployment-oriented modifications: Grouped-Query
Attention~\cite{ainslie2023gqa} with only 4 KV heads, Sliding Window
Attention~\cite{beltagy2020longformer} on three of every four layers, and a
single Multi-Token Prediction (MTP)~\cite{gloeckle2024mtp} head that is used both as an auxiliary pre-training objective and as a built-in draft for speculative decoding.

Our key contributions are:

\begin{itemize}
  \item \textbf{An efficiency-aware architecture.} We systematically ablate
    dense versus MoE backbones, Grouped-Query Attention configurations,
    Multi-head Latent Attention~\cite{deepseekai2024deepseekv2}, Sliding
    Window Attention patterns, and expert sparsity ratios. The resulting 12B/2.5B-active configuration matches or exceeds the throughput of a 7B dense baseline while occupying a substantially larger total-parameter envelope.
  \item \textbf{A three-phase pre-training curriculum on ${\sim}$10.6T tokens.}
    Following the ``web early, curated late''
    paradigm~\cite{grattafiori2024llama3}, the data mixture progressively
    shifts from diverse web content toward curated code and mathematical
    content (code ratio 23\,\%\,$\to$\,42\,\%\,$\to$\,59\,\%), with batch-size
    doubling and an extended capability-sharpening phase that decays the
    learning rate linearly to zero.
  \item \textbf{A Muon + FP8 training recipe at production scale.} We adopt
    the Muon optimizer~\cite{jordan2024muon,liu2025muon} for large-scale MoE
    pre-training, combine it with FP8 hybrid mixed
    precision~\cite{micikevicius2022fp8} and a Warmup-Hold-Decay
    schedule~\cite{hu2024minicpm,hagele2024scaling} with linear decay to
    zero, and report training-stability observations across the full
    ten-trillion-token run.
  \item \textbf{Long-context extension to 128K.} We extend the pre-trained
    base to 131{,}072 tokens following the layer-selective scaling recipe
    of Gemma~3~\cite{team2025gemma3} and OLMo~3~\cite{olmo3} with
    YaRN~\cite{peng2024yarn} as the scaling method, and report empirical
    findings on data-mix transfer and MoE router dynamics during this
    stage.
  \item \textbf{Two post-trained variants from a shared base.} From the same
    long-context checkpoint we produce an \emph{Instruct} model that answers
    directly and a \emph{Thinking} model that emits an explicit reasoning
    trace, each refined further by reinforcement learning with verifiable
    rewards (RLVR) on math and executable coding tasks.
  \item \textbf{Open release.} We release base, instruct, and thinking
    checkpoints under the Apache~2.0 license, together with this report
    documenting the architecture decisions, data pipeline, and training
    recipe behind them. In addition, we release a base model before the long context extension and SFT checkpoints.
\end{itemize}

Across a panel of code generation, math and reasoning, tool use, knowledge,
and safety benchmarks, \modelname is competitive with open-weight baselines
in the 4--14B range despite running at the per-token compute of a 2.5B dense
model, and matches or exceeds the inference throughput of Qwen2.5-7B~\cite{qwen2.5} on a
single H100. The remainder of this report follows the contributions above:
\Cref{sec:architecture} traces the architecture design and ablations,
\Cref{sec:pretraining} details the pre-training data and recipe,
\Cref{sec:long-context} describes the 128K context extension,
\Cref{sec:post-training} covers SFT, RL, and post-training evaluation, and
\Cref{sec:efficiency} reports our inference benchmarks.

\section{Model Architecture}
\label{sec:architecture}

\modelname is a decoder-only Transformer that closely follows the Qwen3-MoE
recipe~\cite{yang2025qwen3}: a Mixture-of-Experts (MoE) feed-forward network in
every layer, Grouped-Query Attention (GQA)~\cite{ainslie2023gqa} with QK-Norm~\cite{henry2020qknorm},
SiLU-gated MLPs~\cite{shazeer2020glu}, RMSNorm~\cite{zhang2019rmsnorm}, and Rotary Position Embeddings (RoPE)~\cite{su2024roformer}.
On top of this backbone we add two latency- and quality-oriented modifications:
Sliding Window Attention (SWA) on a fraction of the layers, and a single
Multi-Token Prediction (MTP) head trained as an auxiliary objective.

\subsection{Architecture Design Decisions}
\label{sec:arch-decisions}

As \modelname is meant to be deployed in JetBrains IDEs, we approached the design space
from the perspective of efficient inference. We targeted the latency and throughput 
budget of a Qwen2.5-7B dense model on a single H100 GPU as our baseline and conducted
several architectural ablations to match it.

\subsubsection{Dense vs.\ Sparse}

We first evaluated whether a dense architecture could outperform the baseline
under our latency constraint. We explored multiple Qwen3-based dense
configurations---varying depth (24--40 layers) and width (hidden sizes 2304--4096),
as well as DeepSeek-style models with Multi-head Latent Attention
(MLA)~\cite{deepseekai2024deepseekv2}. None of the dense configurations
consistently outperformed Qwen2.5-7B on our evaluation benchmarks within the
latency budget. MLA allowed scaling to approximately 5.5B parameters at
equivalent speed, but the quality gains were insufficient to justify the
additional training complexity, and the supported latent rank was too large
for our model scale.

We therefore adopted a Mixture-of-Experts (MoE) architecture, which enabled scaling 
to ${\sim}$12B total parameters while keeping the per-token compute comparable to a 2.5B
dense model.

\subsubsection{Expert Configuration}

Starting from the Qwen3-30B-A3B architecture~\cite{yang2025qwen3}, we scaled
down the model proportionally to fit within a single H100 GPU ($<$18B total
parameters). We fixed the number of experts at 64 as larger expert counts
exceeded GPU memory constraints.

\begin{figure}[H]
  \centering
  \begin{subfigure}[t]{0.48\textwidth}
    \centering
    \includegraphics[width=\textwidth]{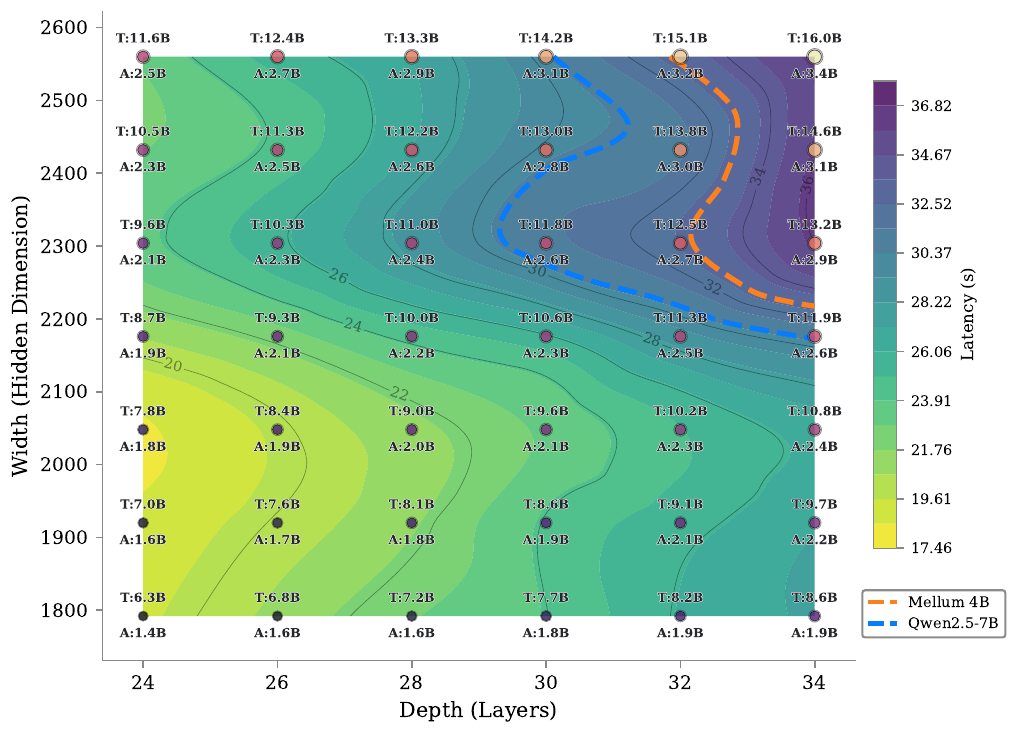}
    \caption{Throughput mode.}
    \label{fig:moe-throughput}
  \end{subfigure}
  \hfill
  \begin{subfigure}[t]{0.48\textwidth}
    \centering
    \includegraphics[width=\textwidth]{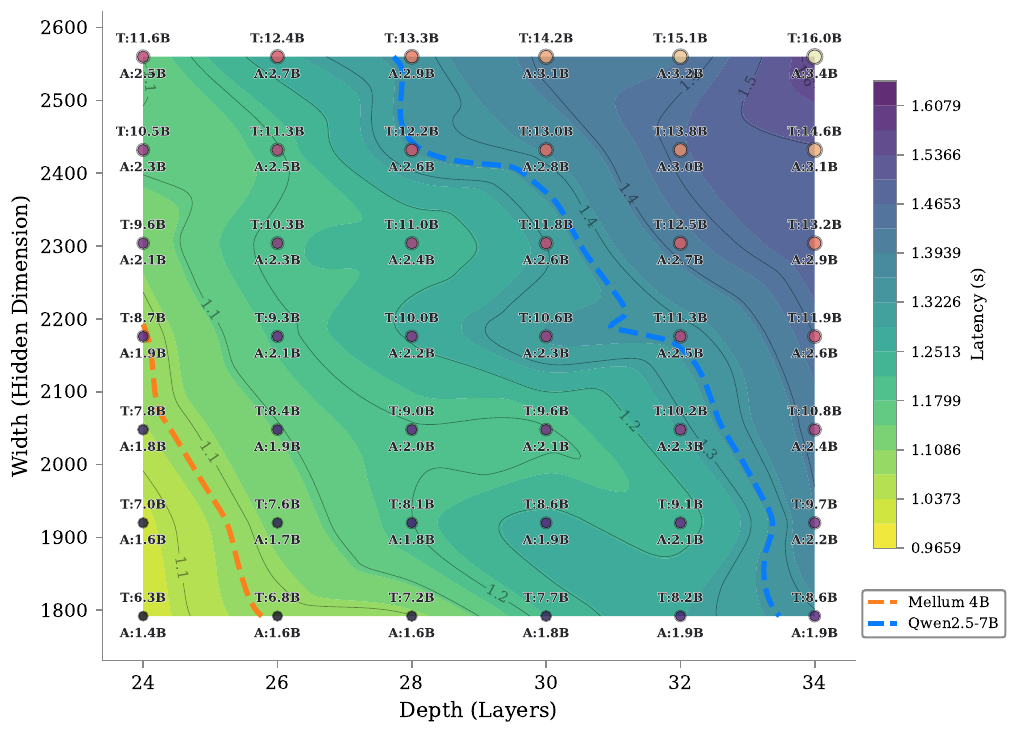}
    \caption{Sync mode.}
    \label{fig:moe-sync}
  \end{subfigure}
  \caption{Iso-latency maps for Qwen3-MoE architectures (64 experts, 8 active)
    across different hidden dimensions and layer counts. Each grid point
    is labelled with \emph{T} (total parameters) and \emph{A} (active
    parameters), both in billions. Dashed lines show the latency
    contours of Mellum~4B (orange) and Qwen2.5-7B (blue); configurations
    below these lines are faster than the corresponding reference model.}
  \label{fig:moe-latency}
\end{figure}

We evaluated different sparsity levels (number of active experts) and found that
higher sparsity (fewer active experts) yielded better inference performance. For
example, 2 active experts achieved ${\sim}$1.5$\times$ lower latency than 8
active experts. However, consistent with prior work suggesting that high
sparsity can be detrimental at smaller scales~\cite{clark2022unified,krajewski2024scaling}, our benchmark evaluations
confirmed that models with lower sparsity (more active experts) produced
better quality. We settled on \textbf{8 active out of 64 total experts} as
the optimal quality--latency trade-off. Under this configuration, the model
supports up to ${\sim}$15B total parameters while matching Qwen2.5-7B latency.
\Cref{fig:moe-latency} shows iso-latency maps for MoE configurations with
8 active experts, illustrating the feasible design space.

\subsubsection{Grouped-Query Attention}

The number of KV heads is the most significant factor affecting inference
throughput under high-concurrency conditions. While the effect is negligible
in synchronous (single-request) mode where KV-cache utilization is low, it
becomes substantial in throughput-dominant serving scenarios. For instance, Qwen2.5-7B
with 4 KV heads achieves roughly the same throughput as our predecessor
Mellum-4B with 8 KV heads despite being nearly twice the size.

\begin{figure}[H]
  \centering
  \begin{subfigure}[t]{0.48\textwidth}
    \centering
    \includegraphics[width=\textwidth]{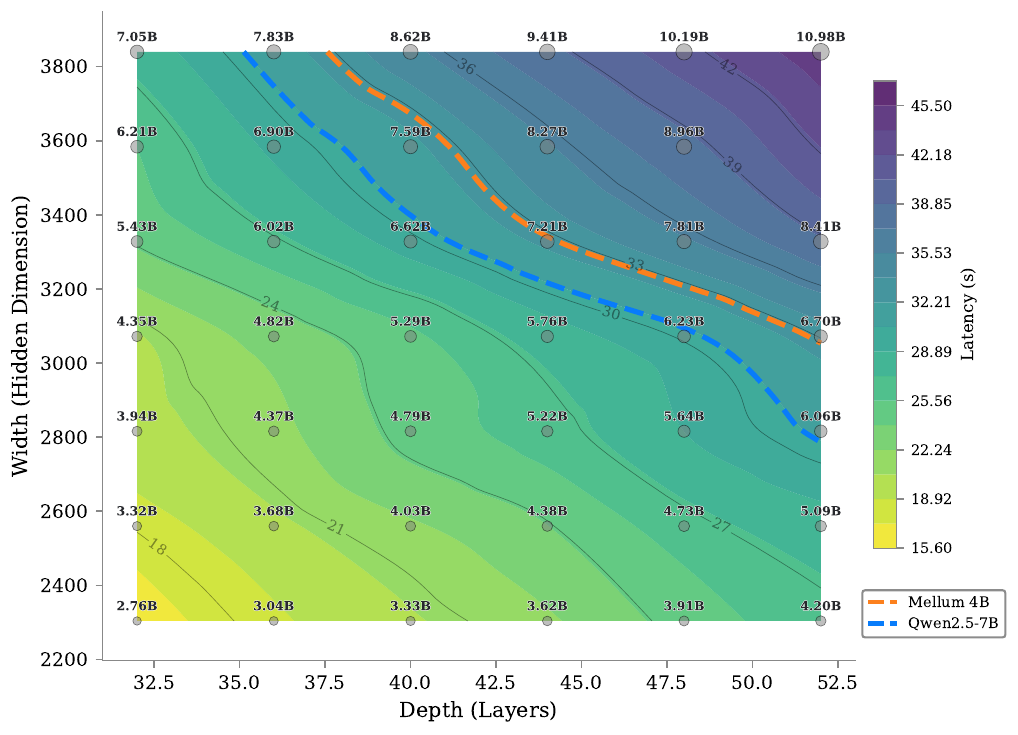}
    \caption{Throughput mode (concurrent requests).}
    \label{fig:kv-heads-throughput}
  \end{subfigure}
  \hfill
  \begin{subfigure}[t]{0.48\textwidth}
    \centering
    \includegraphics[width=\textwidth]{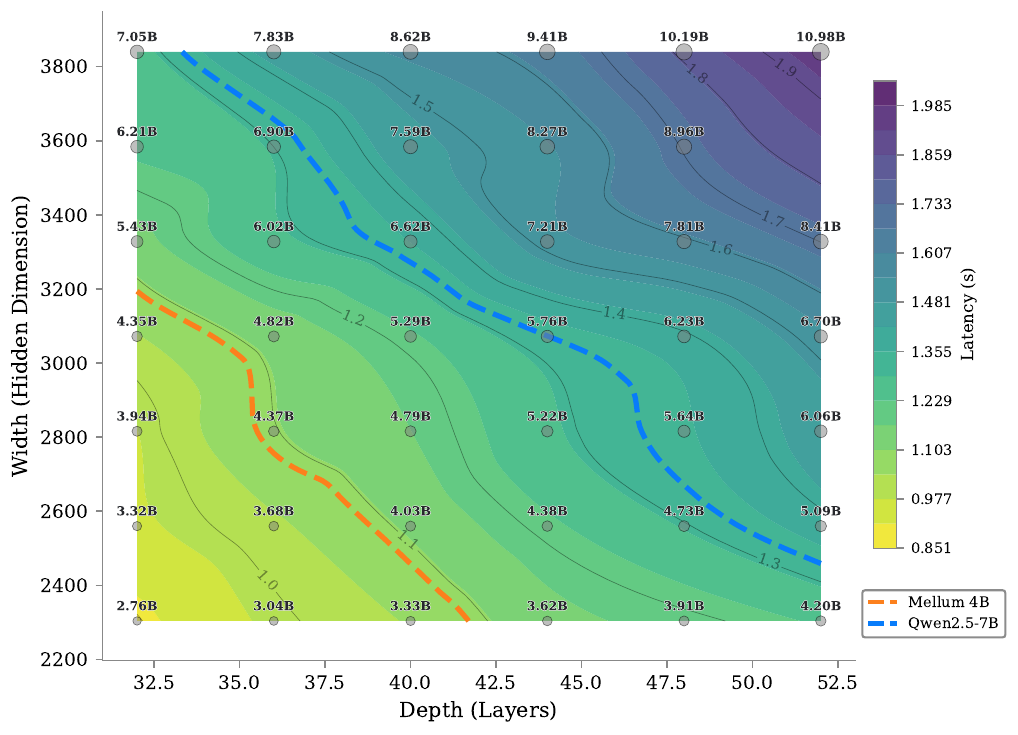}
    \caption{Sync mode (sequential requests).}
    \label{fig:kv-heads-sync}
  \end{subfigure}
  \caption{Iso-latency maps for dense Qwen3 architectures with 4 KV
    heads. Each grid point is labelled with the model's total parameter
    count in billions (e.g., \emph{4.20B}); circle size encodes the
    same quantity. Dashed lines show the latency contours of
    Mellum~4B (orange) and Qwen2.5-7B (blue); configurations below
    these lines are faster than the corresponding reference model.}
  \label{fig:kv-heads}
\end{figure}

We selected \textbf{4 KV heads} as the optimal trade-off: 8 heads caused
significant throughput degradation, while 2 heads yielded insufficient quality
on evaluation benchmarks. \Cref{fig:kv-heads} shows iso-latency maps for
Qwen3-based dense architectures with 4 KV heads, with dashed lines indicating
the latency of Mellum~4B and Qwen2.5-7B. In throughput mode, the KV-cache
bottleneck is clearly visible: wider models (larger hidden dimension) are
disproportionately penalized. In sync mode, where the KV cache is underutilized,
the effect is much smaller and latency is dominated by model depth.

\subsubsection{Sliding Window Attention}

\begin{figure}[H]
  \centering
  \includegraphics[width=0.78\textwidth,height=0.22\textheight,keepaspectratio]{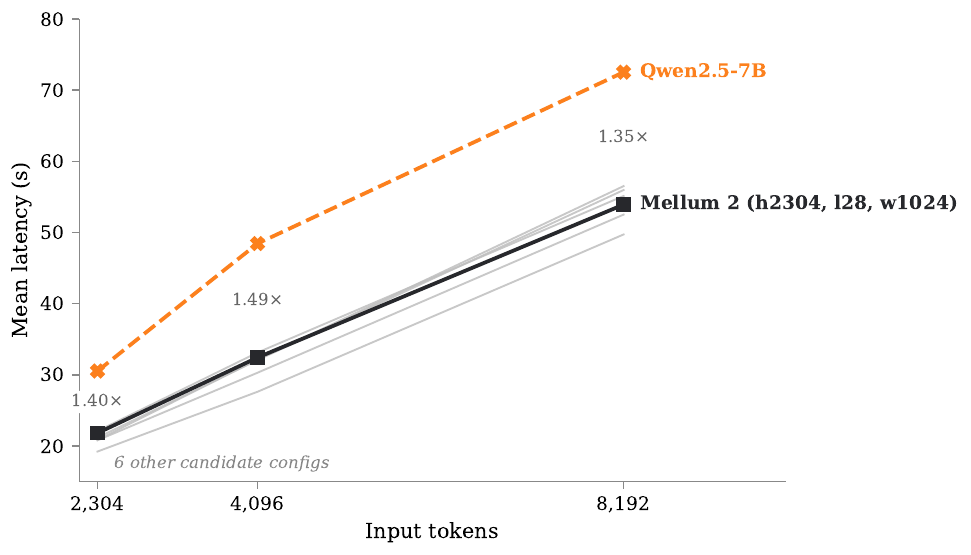}
  \caption{Latency comparison of MoE models with Sliding Window Attention
    (window sizes 512 and 1{,}024, applied to 3/4 of attention layers) against 
    Qwen2.5-7B across different input lengths.}
  \label{fig:swa-latency}
\end{figure}

We adopted Sliding Window Attention (SWA)~\cite{jiang2023mistral,beltagy2020longformer}
as a latency optimization. Experiments on both dense and MoE architectures
confirmed that SWA reduces inference latency by limiting the attention span of
most layers. We apply SWA to \textbf{3 out of every 4 layers} (the remaining
layers use full attention) with a window size of 1{,}024 tokens. This pattern
preserves long-range context capability through the full-attention layers while
reducing compute in the majority of layers. Consistent with findings from the
Gemma model family~\cite{team2025gemma3}, a window size of 1{,}024 outperforms one of size 
512 on quality benchmarks. \Cref{fig:swa-latency} shows that MoE models with
SWA achieve latency comparable to Qwen2.5-7B even at double the context length,
providing a significant advantage in workflows requiring larger context.

\subsubsection{Multi-Token Prediction}
\label{sec:mtp}

We augment the standard next-token prediction objective with a Multi-Token
Prediction (MTP) head~\cite{gloeckle2024mtp} that predicts one additional
future token. The MTP head is a single additional transformer layer that
receives the hidden states from the main model and is trained with a scaled
loss ($\alpha = 0.1$). The MTP head is removed during evaluation and inference
(it does not affect the main model's predictions), but provides a natural draft
model for speculative decoding.

In ablation studies involving a 14B MoE model trained on 105B data tokens, MTP yielded
significant benchmark improvements at a cost of only 7\% additional training
time. The validation loss curves of runs with and without MTP head were nearly
identical, suggesting that MTP does not degrade the primary next-token prediction
objective. Rather, benchmark evaluation (\cref{tab:mtp-benchmarks}) reveals
substantial improvements on key tasks: HumanEval +10.4, MMLU +3.6,
MMLU-Pro +3.3, and GSM8K +3.0.

\begin{table}[!tb]
  \centering
  \caption{Benchmark comparison between baseline and MTP models (14B MoE,
    105B tokens).}
  \label{tab:mtp-benchmarks}
  \small
  \begin{tabular}{@{}llccc@{}}
    \toprule
    \tableheadercolor
    \textbf{Benchmark} & \textbf{Metric} & \textbf{Baseline} & \textbf{+ MTP} & \textbf{$\Delta$} \\
    \midrule
    HumanEval         & pass@1           & 20.73          & \textbf{31.10} & +10.37 \\
    HumanEval+        & pass@1           & 18.29          & \textbf{26.22} & +7.93  \\
    MMLU              & Accuracy         & 37.49          & \textbf{41.06} & +3.57  \\
    MMLU-Pro          & Exact match      & 19.07          & \textbf{22.32} & +3.25  \\
    GSM8K             & Exact match      & 30.63          & \textbf{33.59} & +2.96  \\
    BBH               & Exact match      & 35.00          & \textbf{37.74} & +2.74  \\
    Code MMLU         & Accuracy         & 32.64          & \textbf{34.38} & +1.74  \\
    HellaSwag         & Norm.\ accuracy  & 58.85          & \textbf{59.29} & +0.44  \\
    MBPP              & pass@1           & 36.80          & 36.80          &  0.00  \\
    Code MMLU (cont.) & Norm.\ accuracy  & \textbf{41.17} & 40.94          & $-$0.23 \\
    MMLU (cont.)      & Norm.\ accuracy  & \textbf{38.70} & 38.29          & $-$0.41 \\
    MBPP+             & pass@1           & \textbf{56.61} & 56.08          & $-$0.53 \\
    ARC-Challenge     & Norm.\ accuracy  & \textbf{41.30} & 40.10          & $-$1.20 \\
    Winogrande        & Accuracy         & \textbf{59.51} & 58.01          & $-$1.50 \\
    \bottomrule
  \end{tabular}
\end{table}

\subsection{Final Architecture}
\label{sec:final-arch}

Bringing the design decisions together, we cast \modelname as a Qwen3-MoE-style
decoder-only Transformer with the following components:

\enlargethispage{2\baselineskip}
\begin{itemize}[itemsep=2pt,topsep=4pt,parsep=0pt]
  \item \textbf{Backbone:} 28 transformer layers, hidden dimension 2{,}304, with
    pre-RMSNorm~\cite{zhang2019rmsnorm} ($\epsilon = 10^{-6}$) and SiLU-gated MLPs~\cite{shazeer2020glu}.
  \item \textbf{Attention:} 32 query heads and 4 KV heads (GQA~\cite{ainslie2023gqa}, head
    dimension 128), QK-Norm~\cite{henry2020qknorm} applied to the query and key projections, and RoPE~\cite{su2024roformer}
    with base $\theta = 500{,}000$.
  \item \textbf{Sliding window attention:} a 3:1 SWA~\cite{beltagy2020longformer} pattern in which 3 out of
    every 4 layers use a sliding window of 1{,}024 tokens and the remaining
    layer uses full attention.
  \item \textbf{Mixture-of-Experts:} 64 routed experts per layer with 8 active
    per token (top-8 routing), expert intermediate size 896, and no shared
    expert.
  \item \textbf{Multi-Token Prediction:} a single MTP~\cite{gloeckle2024mtp} transformer layer trained
    with loss weight $\alpha = 0.1$, used as a draft model for speculative
    decoding~\cite{leviathan2023speculative} and removed at evaluation time.
  \item \textbf{Embeddings:} untied input/output embeddings over a 98{,}304-token
    vocabulary; native context length 8{,}192 tokens (extended to 131{,}072 in
    long-context training, see \cref{sec:long-context}).
\end{itemize}

This configuration totals ${\sim}$12B parameters with ${\sim}$2.5B active per
token.\footnote{All matrix dimensions---hidden size 2{,}304, head dimension 128,
expert intermediate size 896---are kept divisible by 128 or higher powers of
two; violations of this alignment can cost up to a 2$\times$ slowdown in GPU
kernel execution, so the constraint was treated as binding throughout the
search.}
\Cref{tab:model-config} summarizes the full set of hyperparameters.

\begin{table}[!tb]
  \centering
  \caption{Architecture configuration of \modelname.}
  \label{tab:model-config}
  \small
  \setlength{\tabcolsep}{4pt}
  \begin{minipage}[t]{0.48\textwidth}
    \centering
    \begin{tabular}{@{}ll@{}}
      \toprule
      \tableheadercolor
      \multicolumn{2}{@{}l}{\textbf{\textsc{Scale}}} \\
      \midrule
      Total parameters    & ${\sim}$12B \\
      Active parameters   & ${\sim}$2.5B \\
      Vocabulary size     & 98{,}304 \\
      Context length      & 8{,}192\,/\,131{,}072$^{\star}$ \\
      Tied embeddings     & No \\
      \addlinespace[4pt]
      \toprule
      \tableheadercolor
      \multicolumn{2}{@{}l}{\textbf{\textsc{Backbone}}} \\
      \midrule
      Layers              & 28 \\
      Hidden dimension    & 2{,}304 \\
      Activation          & SiLU (gated) \\
      Normalization       & RMSNorm ($\epsilon{=}10^{-6}$) \\
      Position encoding   & RoPE ($\theta{=}500{,}000$) \\
      \bottomrule
    \end{tabular}
  \end{minipage}\hfill
  \begin{minipage}[t]{0.48\textwidth}
    \centering
    \begin{tabular}{@{}ll@{}}
      \toprule
      \tableheadercolor
      \multicolumn{2}{@{}l}{\textbf{\textsc{Attention}}} \\
      \midrule
      Query heads         & 32 \\
      KV heads (GQA)      & 4 \\
      Head dimension      & 128 \\
      QK-Norm             & Yes (RMSNorm) \\
      Sliding window      & 1{,}024 (3:1 SWA) \\
      \addlinespace[4pt]
      \toprule
      \tableheadercolor
      \multicolumn{2}{@{}l}{\textbf{\textsc{Mixture-of-Experts \& MTP}}} \\
      \midrule
      Experts (total)     & 64 \\
      Experts (active)    & 8 (top-8) \\
      Expert MLP size     & 896 \\
      Shared expert       & None \\
      MTP layers          & 1 ($\alpha{=}0.1$) \\
      \bottomrule
    \end{tabular}
  \end{minipage}

  \vspace{4pt}
  {\footnotesize $^{\star}$After the long-context extension stage
  (\cref{sec:long-context}).}
\end{table}

\section{Pre-Training}
\label{sec:pretraining}

\subsection{Training Data}
\label{sec:data}

Our pre-training corpus comprises approximately 10.6 trillion tokens drawn from
diverse sources. We organize the data into three broad categories: web and
general knowledge, source code, and mathematical content.

\subsubsection{Source Code}

The code portion of our corpus includes raw, permissively licensed source code files
collected from public repositories and deduplicated at the file level, web pages
containing code extracted from Common Crawl, and a suite of synthetic and derived code
datasets. The derived datasets augment raw code with natural language
annotations---including code summarizations, functionality extensions,
translations between programming languages, test generation, commit messages,
and task descriptions. We also include synthetic code datasets covering
question answering, code rewriting, code review, transpilation, and
educational explanations. Consistent with ~\cite{hui2024qwen25coder}, we find that
synthetic code data can effectively complement raw code, particularly for
smaller-scale MoE models where data diversity is crucial.

\subsubsection{Web and General Knowledge}

The web data component includes large-scale synthetic web corpora derived
from Common Crawl~\cite{su2024nemotroncc}, educational web
content~\cite{penedo2024fineweb}, educational PDFs, multilingual reasoning
and QA datasets, and curated knowledge sources including SFT data, STEM
instruction data, rewrites of Wikipedia pages, and synthetically generated encyclopedic articles.

\subsubsection{Mathematical Data}

Mathematical data includes math-focused SFT data, math-oriented web content at
multiple quality tiers, permissively licensed math textbooks, and math instruction-tuning data.

\subsubsection{Tokenizer}

We use a custom tokenizer with a vocabulary size of 98{,}304 tokens, identical
to the tokenizer used in Mellum-4B~\cite{mellum2025}. The vocabulary is
designed to provide strong coverage of programming language tokens and
technical terminology.

\subsection{Three-Phase Training Curriculum}
\label{sec:curriculum}

Following the ``web early, curated late'' paradigm established by
Llama~3.1~\cite{grattafiori2024llama3}, DeepSeek-V3~\cite{deepseekai2025deepseekv3},
and SmolLM2~\cite{allal2025smollm2}, and most recently adopted by
Arcee Trinity~\cite{singh2026arcee}, we organize pre-training into three
phases that progressively shift from diverse web content toward high-quality
code and mathematical data. The phase boundaries are aligned with the
Warmup-Hold-Decay (WHD) learning rate schedule~\cite{hu2024minicpm,hagele2024scaling}.

\begin{table}[H]
  \centering
  \caption{Three-phase pre-training curriculum. The data mix progressively
    shifts toward code and math as training progresses.}
  \label{tab:curriculum}
  \small
  \setlength{\tabcolsep}{6pt}
  \begin{tabular}{@{}l
                  S[table-format=2.2]
                  S[table-format=3.1]
                  l
                  S[table-format=2.0]
                  S[table-format=2.0]
                  S[table-format=2.0]@{}}
    \toprule
    \tableheadercolor
    \textbf{Phase} & {\textbf{Tokens (T)}} & {\textbf{\% Total}} & \textbf{LR State} & {\textbf{Web \%}} & {\textbf{Code \%}} & {\textbf{Math \%}} \\
    \midrule
    1: Foundation            &  6.18 & 58.0 & Warmup\,$\to$\,Hold & 70 & 23 &  6 \\
    2: Quality Uplift        &  2.79 & 26.2 & Hold                & 44 & 42 & 14 \\
    3: Capability Sharpening &  1.69 & 15.9 & Decay               & 23 & 59 & 18 \\
    \midrule
    \textbf{Total}           & \bfseries 10.65 & \bfseries 100.0 & & & & \\
    \bottomrule
  \end{tabular}
\end{table}

\textbf{Phase 1: Foundation Building} (${\sim}$6.18T tokens, 58\%). The first phase
establishes broad linguistic capabilities and foundational code understanding
using predominantly web data. The mix is approximately 70\% web and general
knowledge, 23\% code, and 6\% math. This phase covers the learning rate warmup
and the beginning of the hold period.

\textbf{Phase 2: Quality Uplift} (${\sim}$2.79T tokens, 26.2\%). The second phase
shifts toward higher-quality data, with significant code upsampling to 42\%.
High-quality curated datasets, including SFT data, reasoning QA, STEM
instruction data, and knowledge-aligned articles, are introduced in this
phase rather than Phase~1, as curated data is more effective during stable
learning rate than during warmup. New synthetic code datasets covering
question answering, code rewriting, and educational explanations are added.
The raw code corpus enters its second epoch.

\textbf{Phase 3: Capability Sharpening} (${\sim}$1.69T tokens, 15.9\%). The final
phase maximizes coding and mathematical capability during learning rate decay,
when the model is most sensitive to data quality. Code reaches 59\% of the mix.
Additional synthetic code datasets covering code review and cross-language
transpilation are introduced. The raw code corpus enters its third epoch.
Web content is reduced to only the highest-quality curated sources.

\subsubsection{Data Repetition Strategy}

High-quality data is scarce, so we repeat it. Small curated code datasets
(summarization, test generation, translation, commit messages, algorithmic
solutions) are shown across all three phases, and the raw code corpus is seen
for three epochs, contributing roughly 958B tokens. No dataset is repeated
more than 4$\times$ over the full run, which we find to be the point where
further repetition stops yielding gains. Repetition is particularly valuable
for MoE training: high-quality data seen multiple times sharpens expert
specialization in a way that a single pass over noisier data does not.

\subsubsection{Fill-in-the-Middle Objective}
\label{sec:fim}

In addition to standard left-to-right next-token prediction, we train
\modelname with a Fill-in-the-Middle (FIM) objective~\cite{bavarian2022fim},
which is essential for in-IDE code completion where the model must condition
on both the prefix and the suffix of the current cursor position. Documents
selected for FIM are split into a (prefix, middle, suffix) triple at two
uniformly sampled positions and reformatted with sentinel tokens. We use a
50/50 split between the Prefix--Suffix--Middle (PSM) and
Suffix--Prefix--Middle (SPM) orderings in all phases.

The fraction of training documents transformed into FIM examples varies
across the curriculum to match the data composition of each phase. In
Phase~1, the FIM rate is 50\% and is applied to all data, exposing the
model to bidirectional context early when the mix is dominated by web and
general-knowledge text. In Phase~2, the FIM rate is reduced to 10\% so
that the high-quality curated code, reasoning, and instruction data
introduced in this phase is consumed primarily under the standard
left-to-right objective. In Phase~3, the FIM rate is restored to 50\%, but
the transformation is restricted to source-code files only; non-code data
(curated web, math, reasoning) continues to be trained with next-token
prediction. This schedule concentrates FIM training on the data
distribution that most closely matches the downstream completion setting,
while preserving generative quality on natural-language inputs.

\subsection{Quality Filtering and Deduplication}

We apply a multi-stage quality filtering pipeline to the raw data:

\begin{enumerate}
  \item \textbf{Heuristic filtering.} We apply checks on line length, entropy, comment ratio,
    and AST parseability checks for code data. We filter samples with fewer
    than 82 unique tokens (1\% of context size) to eliminate degenerate
    sequences with abnormally low lexical diversity, which we identify
    as a source of periodic training loss drops.
  \item \textbf{Classifier-based filtering.} Quality classifiers at multiple
    tiers are used to stratify web data by quality, enabling phase-appropriate
    data selection.
  \item \textbf{Deduplication.} MinHash-based near-deduplication~\cite{lee2022deduplicating} at the file
    level for code data. For web data, intra-phase deduplication is applied,
    while cross-phase repetition is intentional and aligned with the curriculum
    design.
\end{enumerate}

\subsection{Training Setup}
\label{sec:training-setup}

\subsubsection{Optimizer}

We use the Muon optimizer~\cite{jordan2024muon} with the distributed
configuration described in Moonlight~\cite{liu2025muon}. Muon applies
orthogonalization-based updates to hidden layers while using Adam for
embedding and output layers.

We compared AdamW~\cite{loshchilov2019adamw} and Muon on both a dense Qwen2.5-7B model and our
Qwen3-MoE-14B architecture, each trained for 105B tokens. We evaluated two
Muon configurations: Megatron defaults (extra scale factor 1.0) and the
Moonlight setup (extra scale factor 0.2).

\begin{figure}[H]
  \centering
  \begin{subfigure}[t]{0.48\textwidth}
    \centering
    \includegraphics[width=\textwidth]{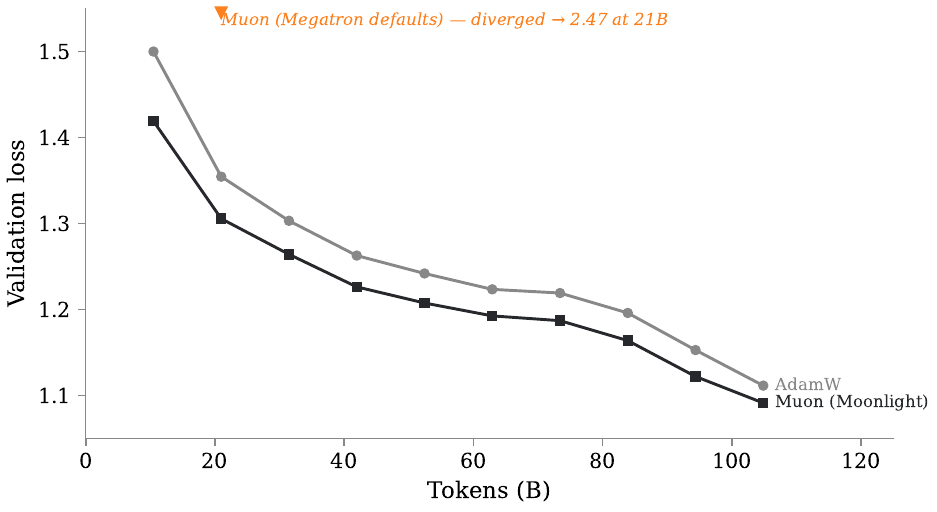}
    \caption{Qwen2.5-7B (dense).}
    \label{fig:muon-qwen25}
  \end{subfigure}
  \hfill
  \begin{subfigure}[t]{0.48\textwidth}
    \centering
    \includegraphics[width=\textwidth]{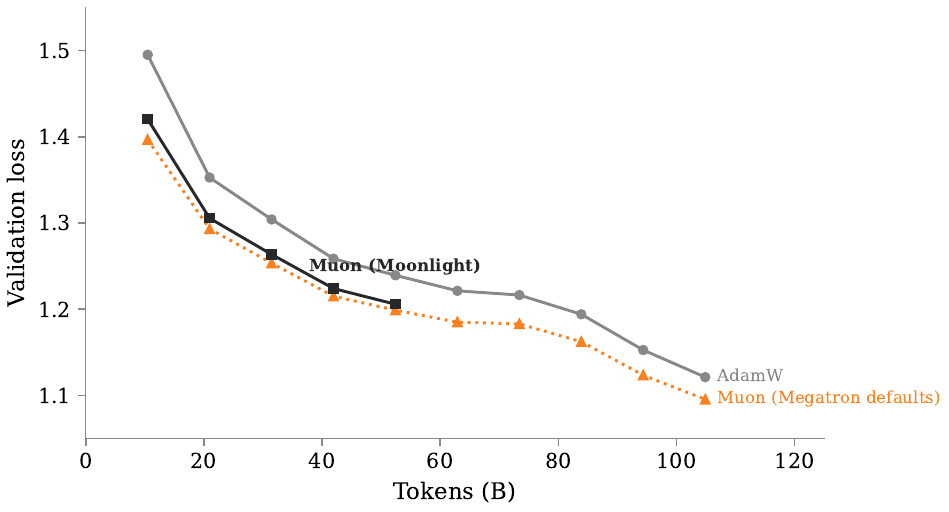}
    \caption{Qwen3-MoE-14B.}
    \label{fig:muon-moe}
  \end{subfigure}
  \caption{Optimizer comparison on 105B-token ablation runs.}
  \label{fig:muon-comparison}
\end{figure}

On the dense 7B architecture (\cref{fig:muon-qwen25}), Megatron defaults caused
immediate divergence, while the Moonlight setup beat AdamW by a large margin,
reducing validation loss by 0.028 (${\sim}$2.5\%). On the MoE-14B
(\cref{fig:muon-moe}), both Muon configurations converged successfully, with
Megatron defaults achieving slightly better final loss ($-$0.026, ${\sim}$2.4\%)
and Moonlight close behind. We selected the Moonlight configuration for its
stability across both dense and MoE architectures.

\begin{table}[H]
  \centering
  \caption{Optimizer and training hyperparameters.}
  \label{tab:hyperparams}
  \small
  \begin{tabular}{@{}ll@{}}
    \toprule
    \tableheadercolor
    \textbf{Hyperparameter} & \textbf{Value} \\
    \midrule
    Optimizer                      & Distributed Muon \\
    Muon momentum                  & 0.95 \\
    Muon Newton--Schulz iterations & 5 \\
    Muon scale mode                & spectral \\
    Muon extra scale factor        & 0.2 \\
    Nesterov momentum              & Yes \\
    Adam $\beta_1, \beta_2$        & 0.9, 0.95 \\
    Adam $\epsilon$                & $10^{-8}$ \\
    Weight decay                   & 0.1 \\
    Gradient clipping              & 1.0 \\
    Peak learning rate             & $3 \times 10^{-4}$ \\
    Minimum learning rate          & 0 \\
    LR schedule                    & WHD (linear decay to zero) \\
    Warmup steps                   & 2{,}000 \\
    Decay steps                    & 49{,}306 \\
    Total training steps           & 323{,}459 \\
    Sequence length                & 8{,}192 \\
    Global batch size (sequences)  & 4{,}096 \\
    Micro batch size               & 2 \\
    Batch rampup                   & 2{,}048 $\to$ 4{,}096 \\
    Precision                      & BF16 + FP8 (hybrid) \\
    \bottomrule
  \end{tabular}
\end{table}

Our investigation of the Adam $\epsilon$ parameter revealed that values as large as
$10^{-5}$ (the value used by LLaMA~2~\cite{touvron2023llama2}) cause disproportionate dampening of
updates. We confirmed that $\epsilon = 10^{-8}$ provides the best trade-off between
training stability and optimization effectiveness.

\subsubsection{Learning Rate Schedule}

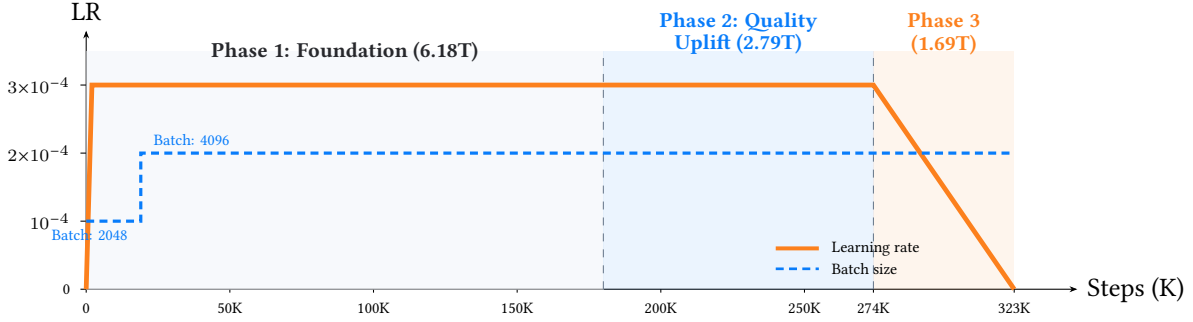
\begin{figure}[H]
\centering
\begin{tikzpicture}[x=0.038cm, y=9cm]
  \draw[-{Stealth[length=3pt]}] (0,0) -- (345,0) node[right] {\small Steps (K)};
  \draw[-{Stealth[length=3pt]}] (0,0) -- (0,0.38) node[above] {\small LR};

  \foreach \y/\label in {0.0/0, 0.1/$10^{-4}$, 0.2/$2{\times}10^{-4}$, 0.3/$3{\times}10^{-4}$} {
    \draw (0,\y) -- (-2,\y) node[left] {\tiny \label};
  }

  \foreach \x/\label in {0/0, 50/50K, 100/100K, 150/150K, 200/200K, 250/250K, 274/274K, 323/323K} {
    \draw (\x,0) -- (\x,-0.005) node[below] {\tiny \label};
  }

  \fill[mellum-light, opacity=0.5] (0,0) rectangle (180,0.35);
  \fill[mellum-mid, opacity=0.08] (180,0) rectangle (274,0.35);
  \fill[mellum-accent, opacity=0.08] (274,0) rectangle (323,0.35);

  \node[above, font=\scriptsize\bfseries, color=mellum-primary] at (90,0.32)
    {Phase 1: Foundation (6.18T)};
  \node[above, font=\scriptsize\bfseries, color=mellum-mid, align=center] at (227,0.33)
    {Phase 2: Quality\\[-1pt] Uplift (2.79T)};
  \node[above, font=\scriptsize\bfseries, color=mellum-accent, align=center] at (298.5,0.33)
    {Phase 3\\[-1pt] (1.69T)};

  \draw[dashed, mellum-gray, thin] (180,0) -- (180,0.35);
  \draw[dashed, mellum-gray, thin] (274,0) -- (274,0.35);

  \draw[mellum-accent, line width=1.8pt]
    (0,0) -- (2,0.3) -- (274,0.3) -- (323,0);

  \draw[mellum-mid, line width=1.2pt, densely dashed]
    (0,0.10) -- (19,0.10) -- (19,0.20) -- (323,0.20);
  \node[font=\tiny, color=mellum-mid, anchor=west] at (20,0.22) {Batch: 4096};
  \node[font=\tiny, color=mellum-mid, anchor=east] at (18,0.08) {Batch: 2048};

  \draw[mellum-accent, line width=1.5pt] (240,0.06) -- (255,0.06);
  \node[font=\tiny, anchor=west] at (256,0.06) {Learning rate};
  \draw[mellum-mid, line width=1.0pt, densely dashed] (240,0.03) -- (255,0.03);
  \node[font=\tiny, anchor=west] at (256,0.03) {Batch size};

\end{tikzpicture}
\caption{Training schedule for \modelname showing the Warmup-Hold-Decay
  (WHD) learning rate schedule, batch size rampup, and three-phase data
  curriculum boundaries.}
\label{fig:lr-schedule}
\end{figure}

We employ a Warmup-Hold-Decay (WHD) schedule~\cite{hu2024minicpm,hagele2024scaling}.
The learning rate warms up linearly over 2{,}000 steps to a peak of $3 \times 10^{-4}$,
holds at peak through Phases~1 and 2, then decays linearly to zero over
Phase~3 (${\sim}$49{,}306 steps, approximately 15\% of total training). The
linear decay to zero follows recent findings showing that it outperforms cosine
decay to a non-zero minimum, providing equivalent loss at lower effective
compute cost. \Cref{fig:lr-schedule} illustrates the full training schedule
with learning rate, batch size rampup, and phase boundaries.

\subsubsection{Batch Size Rampup}

The global batch size ramps linearly from 2{,}048 to 4{,}096 sequences during
the initial phase of training. At full batch size, each step processes
approximately 33.6M tokens ($4{,}096 \times 8{,}192$).

\subsubsection{Precision}
\label{sec:precision}

We use BF16 as the base precision with FP8 hybrid mixed-precision
training~\cite{micikevicius2022fp8}, using tensorwise FP8 recipe with the
most-recent amax algorithm. Gradient reduction is performed in FP32 to maintain
numerical stability.

\subsubsection{MoE-Specific Training}

For the MoE routing, we use global auxiliary load-balancing
loss~\cite{fedus2022switch} with a coefficient of $10^{-3}$, combined with
a router z-loss of $10^{-3}$ for training stability~\cite{zoph2022stmoe}. The router operates in
FP32 precision. We explored both per-sequence and global-batch balancing strategies and
chose global-batch balancing for its flexibility, despite per-sequence balancing producing slightly better loss on short runs.

We adopt \textbf{dropless routing}~\cite{gale2023megablocks} (no expert capacity factor), which avoids
token dropping entirely. In short-run experiments, we found no meaningful
quality difference between capacity factors of 1.0--1.5. Dropless routing
was initially slower than routing with a capacity factor of 1.5 in our tests.
However, this was before accounting for the effect of router balancing on
throughput: as the router learns a proper load balance during training,
dropless routing throughput improves and approaches that of capacity-limited
routing. In the early stages of training, when routing is less balanced, the
overhead is more pronounced. We observe approximately
15\% higher initial iteration step time compared to capacity factor 1.5. Dropless routing
also eliminates information loss from dropped tokens and allows full
micro-batch utilization.

\subsubsection{Sequence Packing}

Documents are combined into fixed-length 8{,}192-token training sequences
using best-fit packing~\cite{ding2024bestfit}, which minimizes
intra-document truncation relative to the standard concatenate-and-chunk
approach and reduces hallucinations caused by spurious cross-document
context.

\subsubsection{Infrastructure}

Training is conducted on 32 nodes, each equipped with 8 H200 GPUs, using a
Megatron-LM~\cite{shoeybi2020megatronlm}-based training framework. We employ
expert parallelism with a degree of 8 (each GPU hosts 8 of 64 experts), with
tensor and pipeline parallelism degrees of 1. Gradient
reduction and parameter gather are overlapped with computation for efficiency.

All data processing is performed offline on a MapReduce-like distributed
storage and compute system. Each raw example is tokenized and then assembled
into fixed-length training-ready shards that are stored alongside the raw
corpora. At training time, a background streamer running on the master node
pulls these shards from the storage cluster and writes them into an in-memory
Redis queue; all data-parallel workers consume batches from this queue over
the internal network. This design fully decouples dataset storage and offline
processing from the training fleet: the two systems share no filesystem and
communicate only through the streaming queue, which lets us place them in
geographically separate data centers (in our setup, the storage and
processing cluster is hosted in France while the training fleet runs on a
GPU cluster in the United States) without exposing transatlantic latency to
the training loop.

\subsection{Training Curves}
\label{sec:training-curves}

\begin{figure}[H]
  \centering
  \begin{subfigure}[t]{0.48\textwidth}
    \centering
    \includegraphics[width=\textwidth]{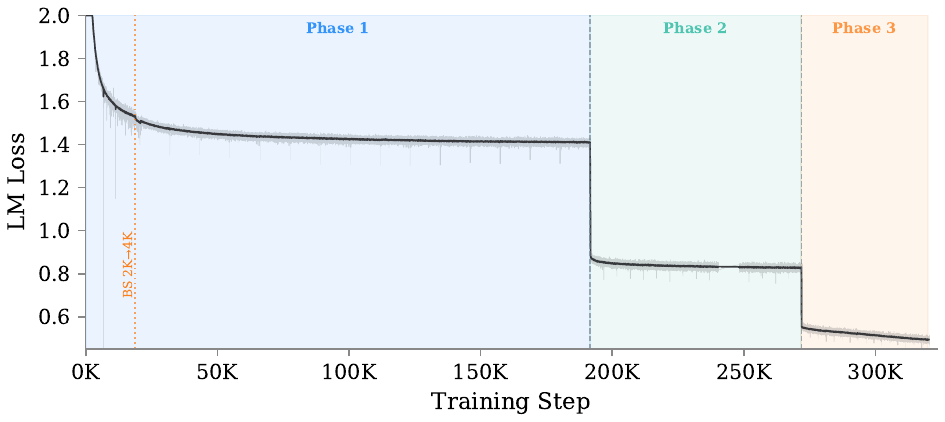}
    \caption{LM loss (next-token prediction).}
    \label{fig:lm-loss}
  \end{subfigure}
  \hfill
  \begin{subfigure}[t]{0.48\textwidth}
    \centering
    \includegraphics[width=\textwidth]{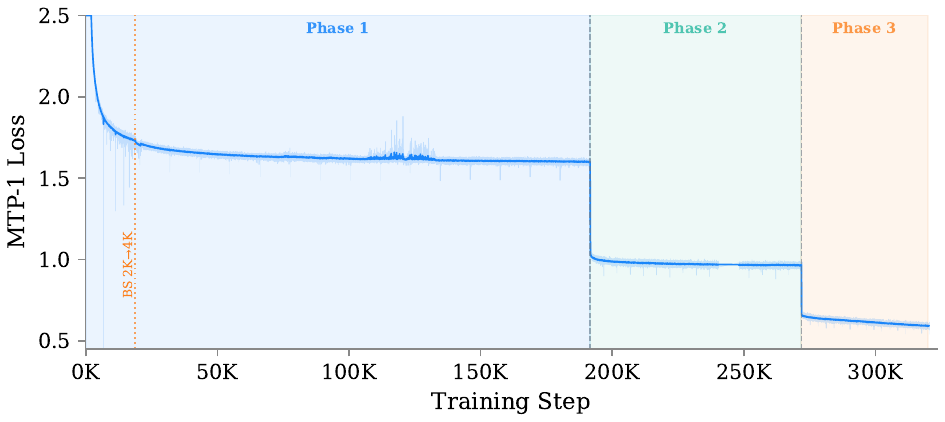}
    \caption{MTP-1 loss (1-step-ahead prediction).}
    \label{fig:mtp-train-loss}
  \end{subfigure}

  \vspace{0.4cm}

  \begin{subfigure}[t]{0.48\textwidth}
    \centering
    \includegraphics[width=\textwidth]{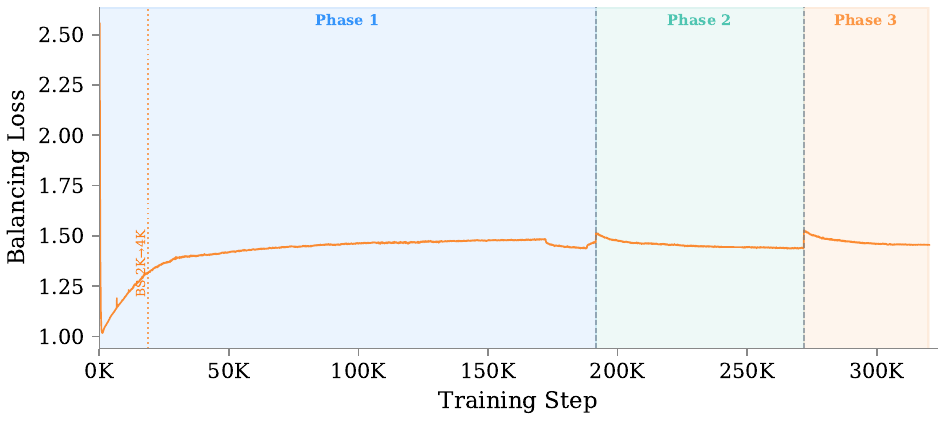}
    \caption{Global load-balancing loss.}
    \label{fig:balancing-loss}
  \end{subfigure}
  \caption{Training loss curves for the \modelname production run. Shaded
    regions indicate the three training phases; the dotted line marks the
    batch size doubling (2{,}048 $\to$ 4{,}096).}
  \label{fig:training-losses}
\end{figure}

\Cref{fig:training-losses} shows the training loss curves from the ongoing
production run. The LM loss decreases steadily across phases, with visible
phase transitions at the data mix boundaries. The MTP loss tracks the LM loss
closely but at a higher magnitude, consistent with the increased difficulty of
predicting tokens further ahead. The global load-balancing loss reflects the
router's learning dynamics: it stabilizes as training progresses, indicating
that the router learns an effective expert assignment.

\subsection{Training Stability}
\label{sec:stability}

During pre-training, we identified and resolved several stability issues:

\textbf{Loss spikes from low-diversity sequences.} Two loss spikes visible
at the very beginning of training were traced to data segments containing
sequences with abnormally low lexical diversity (e.g., a single repeated
token spanning the entire context). We mitigated this by filtering samples
with fewer than 82 unique tokens (1\% of the 8{,}192 context length).

\textbf{Residual periodic loss spikes from hash-sorted duplicates.}
Our data preparation pipeline sorts samples by a hash of the token sequence.
Some source documents were long enough that, when sliced into 8{,}192-token
chunks, multiple chunks became exact duplicates. Hash-based sorting placed
these duplicates at the same position within each data shard. Since each
training phase is composed of 16 uniform shards, the duplicates appear at
roughly the same offset in every shard, producing 16 periodic downward loss
spikes per phase. These spikes are visible in \cref{fig:lm-loss} as faint
periodic dips. We verified that they are modest in magnitude, isolated, and
have no measurable effect on training dynamics---including no impact on the
MoE load-balancing loss (\cref{fig:balancing-loss}). Since removing these
duplicates from the already-prepared data was technically non-trivial, we
chose to continue training with them in place.

\textbf{Cluster migration and load-balancing loss shift.}
Approximately halfway through training, we migrated from 32 nodes to a
smaller cluster of 16 nodes while keeping the effective global batch size
fixed. As visible in \cref{fig:balancing-loss}, the global load-balancing
loss decreased noticeably after this transition. This is not a change in
model behavior but rather a consequence of how Megatron-LM implements the
global auxiliary loss. The implementation maintains a running average of
per-expert token counts across microbatches within each optimizer step,
resetting the accumulator only at gradient finalization. The loss at each
microbatch is computed against this running estimate rather than against a
true global count. When the number of data-parallel ranks changes (here,
halved), the microbatch decomposition of the same effective batch changes:
fewer ranks means more gradient-accumulation microbatches per step, which
allows the running average to converge more closely to the true distribution
before reset. The resulting loss is therefore systematically lower, even
though the effective optimization signal is comparable. This is an
accumulation-semantics artifact rather than a precision issue (all
auxiliary-loss computations use FP32) and did not materially affect
training quality.

\subsection{Pre-Training Evaluation}
\label{sec:pretrain-eval}

We evaluate the base model of \modelname on a broad suite of benchmarks
spanning general knowledge, reasoning, mathematics, and code. We compare against OLMo-3-7B~\cite{olmo3},
Qwen2.5-7B~\cite{qwen2.5}, Qwen3-4B-Base~\cite{yang2025qwen3}, and Qwen3.5-4B-Base~\cite{qwen2026qwen35}.

The evaluation suite consists of 18 benchmarks grouped into three categories:
\begin{itemize}
  \item \textbf{General Knowledge \& Reasoning:} MMLU~\cite{hendrycks2021mmlu},
    MMLU-Pro~\cite{wang2024mmlupro}, BBH~\cite{suzgun2022bbh},
    ARC-Challenge~\cite{clark2018arc}, HellaSwag~\cite{zellers2019hellaswag},
    WinoGrande~\cite{sakaguchi2021winogrande}, and
    TruthfulQA~\cite{lin2022truthfulqa}.
  \item \textbf{Math \& Science:} GSM8K~\cite{cobbe2021gsm8k},
    MATH~\cite{hendrycks2021math}, and GPQA (Main and Diamond
    splits)~\cite{rein2023gpqa}.
  \item \textbf{Code Generation:} HumanEval and HumanEval+~\cite{chen2021humaneval,liu2023evalplus},
    MBPP and MBPP+~\cite{austin2021mbpp,liu2023evalplus},
    MultiPL-E~\cite{cassano2022multipl}, and
    CRUXEval (input and output prediction)~\cite{gu2024cruxeval}.
\end{itemize}

\Cref{tab:pretrain-eval} summarizes performance across all benchmark groups.
Despite activating only 2.5B parameters per token, \modelname is competitive
with 7B dense models on many benchmarks and exceeds them on several reasoning
and code tasks (MMLU-Pro, BBH, GSM8K, MBPP, CRUXEval).

\begin{table}[ht]
\centering
\caption{Pre-training evaluation results. All values are reported as
percentages. The \modelname column is shaded for grouping.}
\label{tab:pretrain-eval}
\small
\setlength{\tabcolsep}{5pt}
\begin{tabular}{@{}l >{\columncolor{mellum-light}}c@{\hspace{12pt}} cccc@{}}
\toprule
\textsc{Benchmark} & \textsc{\modelname} & \textsc{OLMo-3-7B} & \textsc{Qwen2.5-7B} & \textsc{Qwen3-4B} & \textsc{Qwen3.5-4B} \\
 & {\scriptsize\itshape\color{mellum-gray}2.5B/12B} & {\scriptsize\itshape\color{mellum-gray}7B} & {\scriptsize\itshape\color{mellum-gray}7B} & {\scriptsize\itshape\color{mellum-gray}4B} & {\scriptsize\itshape\color{mellum-gray}4B} \\
\midrule
\multicolumn{6}{@{}l}{\textit{\color{mellum-gray}Code Generation}} \\
\addlinespace[2pt]
HumanEval           & 41.5 & 45.1 & 55.5 & 57.3 & 50.0 \\
HumanEval+          & 37.2 & 39.6 & 47.0 & 51.2 & 43.9 \\
MBPP                & 62.4 & 50.6 & 63.6 & 67.0 & 52.2 \\
MBPP+               & 61.4 & 52.9 & 64.0 & 64.5 & 55.0 \\
MultiPL-E (7 langs) & 21.0 & 10.0 & 19.2 & 26.0 & 12.1 \\
CRUXEval-I          & 45.4 & 38.8 & 44.0 & 44.6 & 49.1 \\
CRUXEval-O          & 43.9 & 36.6 & 42.9 & 43.5 & 43.2 \\
\addlinespace[4pt]
\multicolumn{6}{@{}l}{\textit{\color{mellum-gray}Knowledge \& Reasoning}} \\
\addlinespace[2pt]
MMLU                & 70.9 & 62.1 & 71.8 & 71.1 & 74.2 \\
MMLU-Pro            & 59.3 & 34.5 & 48.6 & 51.5 & 52.4 \\
BBH                 & 74.9 & 63.6 & 69.0 & 71.3 & 80.2 \\
ARC-Challenge       & 53.5 & 53.6 & 51.3 & 51.2 & 54.9 \\
HellaSwag           & 73.7 & 74.2 & 78.9 & 73.7 & 75.3 \\
WinoGrande          & 65.5 & 69.5 & 73.3 & 71.2 & 70.8 \\
TruthfulQA MC2      & 44.5 & 47.0 & 56.4 & 53.5 & 52.1 \\
\addlinespace[4pt]
\multicolumn{6}{@{}l}{\textit{\color{mellum-gray}Math \& Science}} \\
\addlinespace[2pt]
GSM8K               & 81.7 & 73.5 & 81.9 & 82.0 & 80.1 \\
MATH                & 10.0 & 18.7 & 24.6 & 27.7 & 25.3 \\
GPQA Diamond        & 31.3 & 28.8 & 32.8 & 36.9 & 41.4 \\
GPQA Main           & 35.0 & 27.9 & 34.2 & 36.8 & 40.2 \\
\bottomrule
\end{tabular}
\end{table}

Key observations:
\begin{itemize}
  \item \textbf{MMLU-Pro}: \modelname achieves 59.3\%, surpassing all comparison
    models including Qwen3.5-4B (52.4\%) and Qwen2.5-7B (48.6\%).
  \item \textbf{BBH}: At 74.9\%, \modelname outperforms OLMo-3-7B (63.6\%),
    Qwen2.5-7B (69.0\%), and Qwen3-4B (71.3\%).
  \item \textbf{GSM8K}: \modelname (81.7\%) is on par with Qwen2.5-7B (81.9\%)
    and Qwen3-4B (82.0\%) despite significantly fewer active parameters.
    \item \textbf{MBPP / MBPP+}: Strong code generation with 62.4\% / 61.4\%,
    outperforming OLMo-3-7B and Qwen3.5-4B.
  \item \textbf{HumanEval}: At 41.5\%, this remains a growth area; we observed significant performance lift on HumanEval after the post-training.
  \item \textbf{GPQA Main}: \modelname (35.0\%) outperforms OLMo-3-7B (27.9\%)
    and Qwen2.5-7B (34.2\%).
\end{itemize}

These results demonstrate that the MoE architecture with 2.5B active parameters
can match or exceed 4--7B dense models on reasoning-heavy benchmarks.

\section{Long Context Extension}
\label{sec:long-context}

Following the main pre-training run, we performed a dedicated long-context
extension stage to extend the effective context length of \modelname from
the 8{,}192-token training context to 131{,}072 tokens (128K).

\subsection{Layer-Selective YaRN}

\begin{figure}[H]
  \centering
  \includegraphics[width=0.72\textwidth,height=0.27\textheight,keepaspectratio]{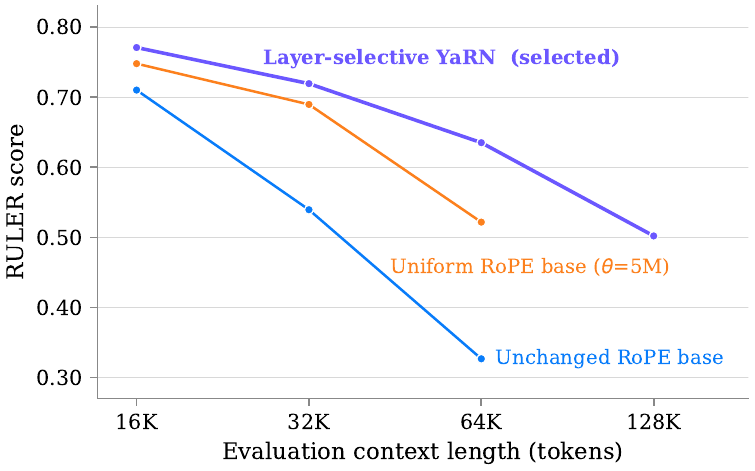}
  \caption{RULER score versus evaluation context length for the three
    long-context recipes we ablated, each scored at its best checkpoint
    along the extension run. The uniform $\theta$-bump and unchanged-$\theta$ evaluation runs were
    capped at a 64K training context, hence the missing 128K points.
    See \cref{app:ruler-formatting} for caveats on the absolute scores.}
  \label{fig:long-context-ablation}
\end{figure}

We adopt YaRN~\cite{peng2024yarn} for context extension, but apply it
selectively rather than uniformly across the network. Specifically, the
YaRN frequency re-mapping is applied only to the global (full-attention)
layers, leaving the sliding window layers with their original RoPE
parameters. This layer-selective recipe was first reported in the
Gemma~3 technical report~\cite{team2025gemma3} (with positional
interpolation rather than YaRN as the scaling method) and was
subsequently adopted by OLMo~3~\cite{olmo3}. Our ablations
(\cref{fig:long-context-ablation}) are consistent with their findings: applying YaRN only to the global layers outperforms both (i) a
uniform RoPE base ($\theta$) bump on all layers and (ii) leaving $\theta$
unchanged. Intuitively, the sliding window layers operate on a fixed
local span and therefore do not require frequency re-mapping, while the
global layers are the only ones that must extrapolate to the new
sequence length.

Concretely, at a 64K evaluation context the layer-selective recipe
reaches a RULER~\cite{hsieh2024ruler} score of $0.64$, compared with $0.52$ for the uniform
$\theta$-bump and $0.33$ for the unchanged-$\theta$ baseline. The gap between recipes \emph{widens} with context
length: the unchanged-$\theta$ run never adapts the full-attention layers
to longer sequences and collapses past 32K, while the uniform bump
unnecessarily perturbs the sliding-window layers that were already
operating correctly at the base context length. The absolute RULER
numbers here are conservative because of a prompt-formatting issue that
depressed scores on the QA subsets throughout the extension stage; we
discuss this in \cref{app:ruler-formatting} and read
\cref{fig:long-context-ablation} as a \emph{within}-recipe ranking
rather than as RULER's final word on absolute long-context capability.

\subsection{Data Mix}

The training data for the extension stage combines a rebalanced version
of the Phase~3 pre-training mix with a portion of agentic SFT data,
which naturally contains long-context examples. The Phase~3 mix was
rebalanced to subsample long reasoning traces, which we found to
dominate the long-context tail and to skew the model toward
reasoning-style outputs at the expense of more general long-context
behaviors.

We also experimented with reproducing OLMo~3's Longmino mix~\cite{olmo3} and several
other mixtures, but were unable to replicate the data-mix gains reported
there. In a head-to-head with everything else held constant (same model,
optimizer, YaRN configuration, and iteration budget), adding the
Longmino mix on top of our base mix \emph{lowered} RULER by roughly
2--3 percentage points at every measured context length, rather than
improving it---consistent with the broader pattern that, across the
configurations we tested, different mixtures produced very similar
benchmark numbers, with our base mix narrowly on top. We also observed
essentially no further quality improvement beyond ${\sim}$30B tokens of
long-context training (\cref{fig:long-context-training}).

To preserve the in-IDE completion capability at long contexts, we also
inject FIM-formatted examples with repository-level context into the
extension mix, following the construction used for \prevmodelname~1~\cite{mellum2025}.
Each example concatenates a set of related files from the same repository
as additional context preceding the (prefix, middle, suffix) target file,
so that the cross-file dependencies relevant to completing the middle span
appear at distances representative of real project layouts. This ensures
that the model learns to attend across repository-scale spans while learning a FIM objective that drives in-IDE completion, similarly to \prevmodelname~1.

\subsection{Training Schedule}

\begin{figure}[H]
  \centering
  \includegraphics[width=0.72\textwidth,height=0.25\textheight,keepaspectratio]{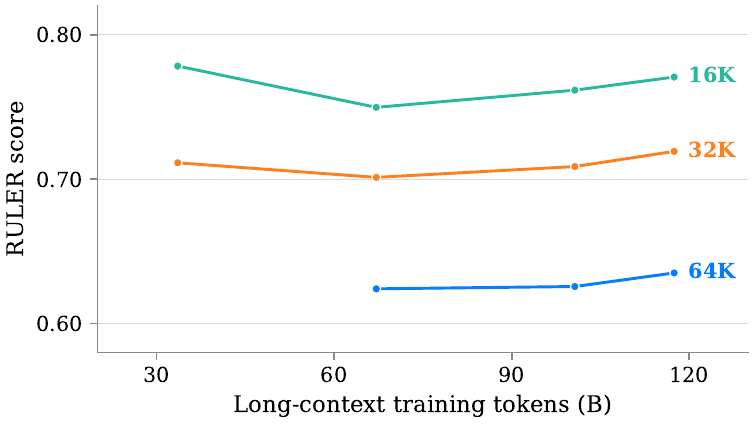}
  \caption{RULER score versus training tokens during the long-context
    extension stage, for the chosen layer-selective YaRN recipe.
    See \cref{app:ruler-formatting} for a comment on absolute RULER scores.}
  \label{fig:long-context-training}
\end{figure}

\begin{figure}[H]
  \centering
  \includegraphics[width=0.72\textwidth,height=0.25\textheight,keepaspectratio]{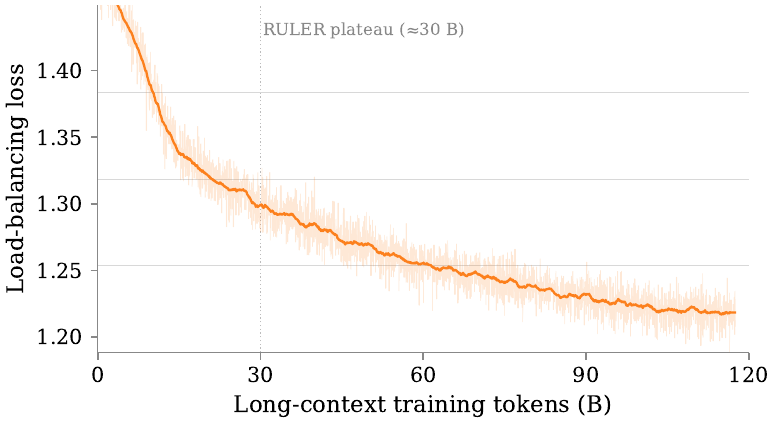}
  \caption{Global MoE load-balancing loss during the long-context
    extension stage.}
  \label{fig:long-context-balancing}
\end{figure}

\Cref{fig:long-context-training} plots RULER scores against the number
of long-context training tokens for the chosen recipe. By the end of
the first ${\sim}$30B tokens, RULER at every measured context length is
already within ${\sim}$1\,pp of the final value reached at 117B tokens;
the subsequent ${\sim}$3$\times$ increase in token budget yields only
marginal improvements. Beyond the 30B-token point, the only quantity
that continued to change meaningfully was the MoE router's
load-balancing loss, which decreased substantially as the router adapted
to the new sequence-length regime (\cref{fig:long-context-balancing}).
On the strength of this signal, we extended the run to 3{,}500 iterations
(${\sim}$117B tokens) using a Warmup-Hold-Decay (WHD)
schedule~\cite{hu2024minicpm,hagele2024scaling} with 500 decay iterations
and a peak learning rate of $3 \times 10^{-5}$, allowing the router to
fully equilibrate before annealing.

\section{Post-Training}
\label{sec:post-training}

Post-training of \modelname starts from the long-context YaRN checkpoint
described in \cref{sec:long-context} and proceeds in two stages: supervised
fine-tuning (SFT) and reinforcement learning.

\subsection{Supervised Fine-Tuning}
\label{sec:sft}

We train two SFT variants of \modelname from the same long-context base
checkpoint and the same data mix, differing in their chat templates and in
how reasoning traces and loss masking are handled:

\begin{itemize}
  \item \textbf{Instruct (no-thinking).} A general-purpose assistant that
    produces answers directly, without an externalized chain of thought.
    Loss is computed on every assistant turn in the conversation, with all
    other tokens masked, and any reasoning fields present in the source data
    are discarded.
  \item \textbf{Thinking.} A reasoning-augmented assistant that emits an
    internal chain of thought before its final answer. Only the final
    assistant turn, together with its reasoning trace, contributes to the
    loss; preceding turns serve as conditioning context, and conversations
    lacking a reasoning trace are excluded. To amplify the effective signal
    on multi-turn data, each multi-turn conversation is unfolded by sliding
    the loss target across successive assistant turns, producing up to five
    training samples per source conversation.
\end{itemize}

After tokenization, sequences are packed to the full 131{,}072-token training
length; samples that would not fit cleanly into a pack are dropped rather
than truncated. Both variants reuse the pre-training optimizer and precision
stack and keep the Multi-Token Prediction head active throughout SFT.

\subsubsection{Data Composition}
\label{sec:sft-data}

The SFT corpus is assembled from a number of sources covering the
capabilities we want \modelname to provide at deployment time. The 
dataset mix can be grouped into the following broad categories:

\begin{itemize}
  \item \textbf{General chat and instruction-following.} Single- and
    multi-turn conversational data covering open-domain questions,
    reading-comprehension QA, multiple-choice items, and short-form
    instruction-following. 
  \item \textbf{Single-turn coding.} Code generation, editing, explanation,
    and translation prompts spanning multiple programming languages, with
    dedicated splits for C++, Python, C\#, JavaScript and TypeScript competitive programming.
  \item \textbf{Agentic coding.} Long-horizon interactive agent trajectories
    (early and revised generations), including SWE-style repository-level edit tasks. These supply
    the model with patterns for navigating a codebase, planning multi-step
    edits, and verifying intermediate results.
  \item \textbf{Tool use and function calling.} Tool-augmented
    conversations covering general function-calling formats, Bash execution, a
    clarification tool, and search tools. The mix
    teaches both schema-faithful tool invocation and recovery from tool
    errors.
  \item \textbf{Reasoning traces.} Chain-of-thought-bearing examples that
    populate the \texttt{reasoning} field used by the thinking variant.
    These cover math, code, and general reasoning; they are filtered out at
    processing time for the instruct variant.
  \item \textbf{Safety.} Refusal and safe-response data drawn from a
    permissively licensed safety corpus, included to reduce harmful
    completions without degrading helpfulness on benign code prompts.
  \item \textbf{Identity examples.} A small set of self-identification dialogues is
    oversampled (3$\times$) so that the model reliably introduces itself as
    \modelname rather than its upstream architectures. Interestingly, in initial runs without identity data, the model consistently identified itself as an AI assistant developed by Google, even though no Google models were used for synthetic data generation.
\end{itemize}

Every example is stored in a unified schema with a \texttt{messages} list
(role/content turns), an optional \texttt{tools} list describing available
function-call signatures, and an optional \texttt{reasoning} field holding
the chain-of-thought associated with the final assistant turn.

\subsubsection{Training Setup}
\label{sec:sft-training}

Both SFT runs initialize from the long-context YaRN checkpoint
(\cref{sec:long-context}), use the same distributed Muon optimizer as
pre-training, and run for three epochs over their respective packed
datasets. The learning rate peaks at $3{\times}10^{-5}$---a tenth of the
pre-training peak---warms up linearly over 100 iterations, and then
decays cosine-style to $3{\times}10^{-6}$ (10\,\% of peak) over the
remainder of training. We keep BF16 with FP8 hybrid mixed precision, the dropless MoE
router, and the MTP head with loss weight $\alpha = 0.1$ unchanged from
pre-training. The MoE auxiliary load-balancing coefficient is reduced from
$10^{-3}$ to $10^{-4}$, since the router is already well-balanced after
pre-training and a smaller coefficient avoids over-constraining expert
utilization on the narrower SFT distribution.

We train at a global batch size of 64 packed sequences of length
131{,}072—roughly 8.4M tokens per optimizer step—on 16 nodes of 8
H200 GPUs each. The run uses expert parallelism of 8 and context
parallelism of 8. The instruct run consumes $\approx$\,47B tokens and the
thinking run $\approx$\,167B tokens, matching the three-epoch budget on
each packed dataset. \Cref{tab:sft-hyperparams} summarizes the shared and
variant-specific hyperparameters.

\begin{table}[H]
  \centering
  \caption{Supervised fine-tuning configuration. Shared rows apply to both
  runs; rows below the rule differ between Instruct and Thinking.}
  \label{tab:sft-hyperparams}
  \small
  \begin{tabular}{@{}lcc@{}}
    \toprule
    \tableheadercolor
    \textbf{Hyperparameter} & \textbf{Instruct} & \textbf{Thinking} \\
    \midrule
    Base checkpoint        & \multicolumn{2}{c}{Long-context YaRN 128K (\cref{sec:long-context})} \\
    Optimizer              & \multicolumn{2}{c}{Distributed Muon (same configuration as pre-training)} \\
    Peak learning rate     & \multicolumn{2}{c}{$3 \times 10^{-5}$} \\
    Minimum learning rate  & \multicolumn{2}{c}{$3 \times 10^{-6}$ (10\,\% of peak)} \\
    LR schedule            & \multicolumn{2}{c}{Cosine decay, 100-iter linear warmup} \\
    Weight decay           & \multicolumn{2}{c}{0.1} \\
    Gradient clipping      & \multicolumn{2}{c}{1.0} \\
    Sequence length        & \multicolumn{2}{c}{131{,}072 (packed)} \\
    Global batch size      & \multicolumn{2}{c}{64 sequences ($\approx$\,8.4M tokens/step)} \\
    Precision              & \multicolumn{2}{c}{BF16 + FP8 hybrid (tensorwise, most-recent amax)} \\
    MTP head               & \multicolumn{2}{c}{Retained, $\alpha = 0.1$} \\
    MoE aux-loss coefficient & \multicolumn{2}{c}{$10^{-4}$} \\
    Expert parallelism     & \multicolumn{2}{c}{8} \\
    Context parallelism    & \multicolumn{2}{c}{8} \\
    Hardware               & \multicolumn{2}{c}{16\,$\times$\,8 H200 GPUs} \\
    Epochs                 & \multicolumn{2}{c}{3} \\    
    \midrule
    Loss mode                 & All assistant turns & Last assistant turn only \\
    Reasoning in target       & No                  & Yes \\
    Turn unfolding            & No                  & Yes (up to 5 samples/conv.) \\
    Packed sequences          & 119{,}334           & 425{,}322 \\
    Training tokens           & $\approx$\,47B      & $\approx$\,167B \\
    \bottomrule
  \end{tabular}
\end{table}

\subsection{Reinforcement Learning}
\label{sec:rl}

Post-training of \modelname finishes with a Reinforcement Learning (RL) stage
that refines each SFT checkpoint against programmatically verifiable
rewards (RLVR). We use RLVR rather than RLHF because every prompt in our
training corpus admits an unambiguous, programmatic correctness check, so
we never have to train a separate reward model whose noise could
dominate the gradient signal.

We run RL twice, once per SFT variant. The \textbf{Instruct} stage starts
from the SFT-instruct checkpoint and trains on the data mix for the
Instruct model. The \textbf{Thinking} stage is a cold restart from the
SFT-thinking checkpoint on the data mix for the Thinking model, and its tasks are more difficult for the model than the Instruct mix because it adds a more challenging long-form math subset. Each stage produces its own deployable checkpoint; the two
runs never share weights.

Both stages use a variation of GRPO~\cite{shao2024deepseekmath} with a
few adjustments that we describe later in this section.

\subsubsection{Infrastructure}
\label{sec:rl-infra}

RL runs on a single Kubernetes cluster of H200 GPU nodes. The cluster is
split into two roles at launch time: a small
group of \emph{training} nodes that owns the policy weights and runs the
gradient updates, and a larger group of \emph{inference} nodes that hosts
the generation engines and produces the rollouts. The split is fixed for
the duration of a run.

\paragraph{Training stack.}
The trainer is built on NeMo-RL~\cite{nemo-rl}, which already provides
the asynchronous GRPO loop we use. Model parallelism, optimizer state, and the policy
backward pass go through Megatron-Bridge, configured with the same MoE
routing, attention layout, and BF16 / FP8 hybrid precision recipe used
during pre-training (\cref{sec:precision}). Generation runs in
vLLM~\cite{kwon2023vllm}. The whole pipeline is orchestrated by Ray and
scheduled by Kubernetes.

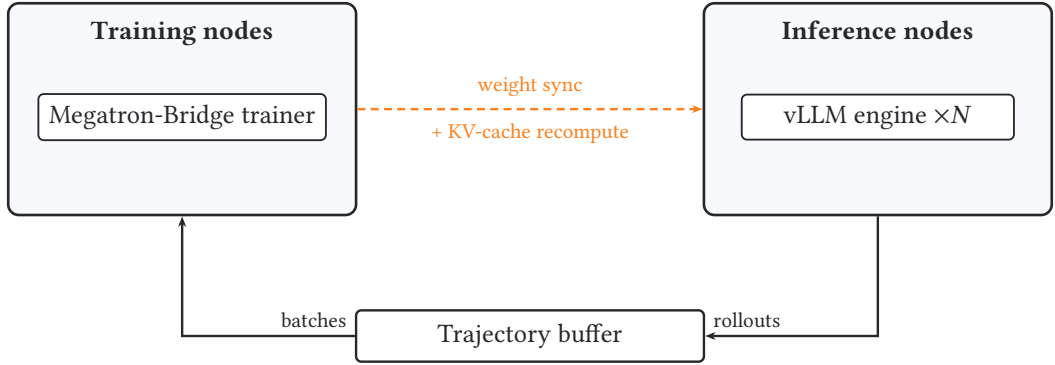
\begin{figure}[H]
\centering
\begin{tikzpicture}[
  font=\small,
  every node/.style={align=center},
  inner/.style={rectangle, draw=mellum-primary, line width=0.7pt, rounded corners=2pt, minimum height=1.5em, inner sep=4pt, fill=white, text=mellum-primary},
  cluster/.style={rectangle, draw=mellum-primary, line width=1.0pt, rounded corners=4pt, inner sep=8pt, fill=mellum-light, fill opacity=0.5, text opacity=1},
  buf/.style={rectangle, draw=mellum-primary, line width=0.9pt, rounded corners=2pt, minimum height=1.7em, inner sep=5pt, fill=white, text=mellum-primary},
  arrow/.style={-{Stealth[length=4pt]}, line width=0.9pt, mellum-primary},
  weight/.style={-{Stealth[length=4pt]}, line width=0.9pt, mellum-accent, densely dashed}
]
  \node[cluster, minimum width=4.6cm, minimum height=2.8cm] (trainer-c) at (0,0) {};
  \node[font=\small\bfseries, color=mellum-primary, anchor=north] at ([yshift=-3pt]trainer-c.north) {Training nodes};
  \node[inner, minimum width=3.6cm, anchor=center] (trainer) at ([yshift=-3pt]trainer-c.center) {Megatron-Bridge trainer};

  \node[cluster, minimum width=4.6cm, minimum height=2.8cm] (infer-c) at (9.2,0) {};
  \node[font=\small\bfseries, color=mellum-primary, anchor=north] at ([yshift=-3pt]infer-c.north) {Inference nodes};
  \node[inner, minimum width=3.6cm, anchor=center] (vllm) at ([yshift=-3pt]infer-c.center) {vLLM engine $\times N$};

  \node[buf, minimum width=4.6cm] (buf) at (4.6,-3.0) {Trajectory buffer};

  \draw[arrow] (infer-c.south) |- (buf.east);
  \node[font=\scriptsize, color=mellum-primary, anchor=south west, inner sep=1pt] at ([xshift=2pt,yshift=2pt]buf.east) {rollouts};

  \draw[arrow] (buf.west) -| (trainer-c.south);
  \node[font=\scriptsize, color=mellum-primary, anchor=south east, inner sep=1pt] at ([xshift=-2pt,yshift=2pt]buf.west) {batches};

  \draw[weight] (trainer-c.east) -- (infer-c.west)
    node[midway, above=1pt, font=\scriptsize, color=mellum-accent] {weight sync}
    node[midway, below=1pt, font=\scriptsize, color=mellum-accent] {+ KV-cache recompute};
\end{tikzpicture}
\caption{Async GRPO actor topology.}
\label{fig:rl-actors}
\end{figure}

\paragraph{Async actor topology.}
\Cref{fig:rl-actors} summarises the actor topology. Trajectory collectors
stream completed rollouts into a global buffer; the trainer pulls
batches from it, runs the GRPO update, and pushes new weights back to
the inference engines. A trajectory may span two consecutive policy
versions, which we cap to a small staleness window. After every weight
push the inference engines recompute the KV cache so that prefix logits
stay consistent with the new policy.

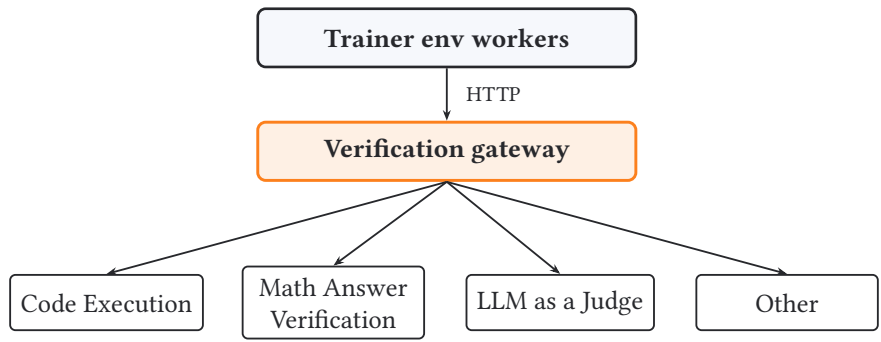
\begin{figure}[H]
\centering
\begin{tikzpicture}[
  font=\small,
  every node/.style={align=center},
  block/.style={rectangle, draw=mellum-primary, line width=0.7pt, rounded corners=2pt, minimum height=1.9em, inner sep=4pt, fill=white, text=mellum-primary},
  gw/.style={rectangle, draw=mellum-accent, line width=1.1pt, rounded corners=3pt, minimum height=2em, inner sep=6pt, fill=mellum-accent, fill opacity=0.12, text opacity=1, text=mellum-primary, font=\small\bfseries},
  trainer/.style={rectangle, draw=mellum-primary, line width=1.0pt, rounded corners=3pt, minimum height=2em, inner sep=6pt, fill=mellum-light, fill opacity=0.6, text opacity=1, font=\small\bfseries, text=mellum-primary},
  arrow/.style={-{Stealth[length=4pt]}, line width=0.7pt, mellum-primary}
]
  \node[trainer, minimum width=5.0cm] (tr) at (0,0) {Trainer env workers};
  \node[gw, minimum width=5.0cm] (gw) at (0,-1.5) {Verification gateway};
  \node[block, minimum width=2.4cm] (code)  at (-4.5,-3.5) {Code Execution};
  \node[block, minimum width=2.4cm] (math)  at (-1.5,-3.5) {Math Answer\\Verification};
  \node[block, minimum width=2.4cm] (judge) at ( 1.5,-3.5) {LLM as a Judge};
  \node[block, minimum width=2.4cm] (other) at ( 4.5,-3.5) {Other};

  \draw[arrow] (tr.south) -- (gw.north) node[midway, right=6pt, font=\scriptsize, inner sep=1pt] {HTTP};
  \foreach \t in {code, math, judge, other} {
    \draw[arrow] (gw.south) -- (\t.north);
  }
\end{tikzpicture}
\caption{Verification stack.}
\label{fig:rl-verifier}
\end{figure}

\paragraph{Verification stack.}
Reward computation is decoupled from the training loop and runs as a
separate set of microservices (\cref{fig:rl-verifier}). The trainer's
environment workers issue HTTP calls into a verification gateway, which
routes each request to the appropriate backend based on the verifier
type carried with each prompt. This decoupling lets us run the entire
verification stack on a separate cluster, so it never competes for GPUs
or memory with the trainer and the generation engines, and it makes
scaling and monitoring each backend independent of the training job.
Backends used during \modelname{} RL include a code execution sandbox
for unit-test based rewards on code, a math answer verifier that
performs symbolic and numeric comparison, an LLM-as-a-Judge service for
grading free-form outputs, and a number of other environments that back
the remaining tasks. Some of those other environments need extra state,
for example session management for stateful tool conversations, so they
sit behind their own dedicated services. The gateway distinguishes
between two kinds of failures during a verification call: the model's
response was un-scoreable, or the verifier itself was transiently
unavailable. We keep these separate so the trainer sees a clean reward
signal: un-scoreable responses produce a zero reward and the model is
shown the error string on its next rollout, while infrastructure
failures are retried.

\subsubsection{Data}
\label{sec:rl-data}

We build two RL data mixes, one per stage. Each is assembled from a
combination of public RLVR releases and a small set of our own
additions, organized into six capability domains: code, math, agentic
tool use, instruction following, reasoning, and knowledge. Both mixes
total roughly 260{,}000 training prompts and 3{,}600 validation prompts,
and \cref{tab:rl-mix} summarises the per-domain breakdown. The two
mixes share most sources and are roughly the same size; the only meaningful difference is that the
Thinking mix replaces part of the pure-math share with a
difficulty-filtered long-form math subset, making it the
harder mix overall.

\begin{table}[H]
  \centering
  \caption{RL data mix composition by capability domain, in number of
  training prompts and share of total.}
  \label{tab:rl-mix}
  \small
  \begin{tabular}{@{}lcccc@{}}
    \toprule
    \tableheadercolor
    \textbf{Domain} & \multicolumn{2}{c}{\textbf{Instruct mix}} & \multicolumn{2}{c}{\textbf{Thinking mix}} \\
    \cmidrule(lr){2-3}\cmidrule(lr){4-5}
                          & Prompts   & Share & Prompts   & Share \\
    \midrule
    Code                  & 57{,}500  & 22\% & 57{,}500  & 22\% \\
    Math                  & 60{,}000  & 23\% & 72{,}000  & 28\% \\
    Agentic tool use      & 36{,}000  & 14\% & 31{,}000  & 12\% \\
    Instruction following & 49{,}500  & 19\% & 53{,}500  & 21\% \\
    Reasoning             & 35{,}000  & 13\% & 35{,}000  & 13\% \\
    Knowledge             & 22{,}500  &  9\% & 10{,}000  &  4\% \\
    \midrule
    Total                 & 260{,}500 & 100\% & 259{,}000 & 100\% \\
    \bottomrule
  \end{tabular}
\end{table}

\paragraph{Code.}
The code domain combines three sources. We use a dataset with
competitive programming problems and tests~\cite{nemo-gym}.
We also use a public math-with-code dataset~\cite{nemo-gym}, which
pairs a hard math prompt with
a Jupyter-style Python execution tool: the model generates Python code, reads
back the tool's stdout, and emits a final answer (this dataset is also
counted under Math in \cref{tab:rl-mix}). On top of these two public
sources, we add our own collection of realistic multi-language coding
tasks covering twelve target languages (Python, Java, PHP,
TypeScript, C\#, JavaScript, JSX, Rust, Kotlin, Go, C++, and CSS) and
grouped by the kind of work the model has to do: greenfield
implementation, debugging from a stack trace, test generation,
behaviour modification, filesystem and API integration, and security
hardening. Each task in this collection ships with a test suite, and
the fraction of passing tests defines the reward signal.

\paragraph{Math.}
Math is the largest single block in both mixes (60{,}000 prompts / 23\% in
Instruct, 72{,}000 prompts / 28\% in Thinking) and is built from three
complementary styles. The first is pure math with no tools, where the
model must do the work in its own context and emit a final answer that
a strict-match verifier compares against the ground truth. For the
Instruct mix we take this subset from the math portion of OLMo-3's
Instruct RL release~\cite{olmo3}; for the Thinking mix we swap in the
math portion of OLMo-3's Thinking RL release~\cite{olmo3}, which is
harder than its Instruct counterpart and the primary contributor to its difficulty. 
The second style is math with calculator
tools, taken from Nemotron's math-advanced-calculations
release~\cite{nemo-gym}, where the model issues calculator-tool calls
and folds the returned values into its answer. The third style is math with code execution, the
math-with-code dataset already described under Code, where the model
uses the Python execution tool to compute intermediate quantities. The
three styles together cover the main ways the deployed model attacks
hard math problems at inference time.

\paragraph{Agentic tool use.}
The math subsets already exercise the tool-use channel, since both the
calculator-tool dataset and math-with-code involve issuing tool calls
and reading back their results. On top of that we add two dedicated
agentic sources. The first is xLAM-style function-calling RLVR
data~\cite{nemo-gym}, where the model picks and parameterises a tool
from an OpenAI-format tool registry in a single step. The second is a
stateful workplace-assistant benchmark~\cite{nemo-gym} in which the
model uses an evolving set of personal-assistant tools (calendar,
email, customer-relations, project-management, and analytics queries)
inside a session-managed environment; the verifier replays the resulting trajectory against a
ground-truth state to score it. These two sources account for 14\% of
the Instruct mix and 12\% of the Thinking mix.

\paragraph{Instruction following.}
The instruction-following block exercises format adherence and
rule-based constraints. We include a generic verifiable IF dataset
graded by machine-checkable instructions, a structured-output dataset
graded by JSON-schema validation, and a small calendar-scheduling
agent, all from Nemotron's public RLVR
release~\cite{nemo-gym}. Together they contribute 19\% of the Instruct
mix and 21\% of the Thinking mix.

\paragraph{Reasoning.}
We include a large slice of \texttt{reasoning-gym}~\cite{stojanovski2025reasoninggym}, a public library of
roughly a hundred procedurally generated reasoning tasks (logic
puzzles, sequence completion, spatial reasoning, simple games) each
with its own task-specific verifier. \texttt{reasoning-gym} keeps the
mix's reasoning footprint broad without committing to any single
benchmark format and contributes about 13\% to both mixes.

\paragraph{Knowledge.}
A multi-domain MCQA pool covers physics, biology, mathematics,
humanities, computer science, engineering, chemistry, and several other
subjects. It is the smallest domain in both mixes (9\% of Instruct,
4\% of Thinking) and is intentionally downsampled because we have
observed that excessive MCQA exposure can hurt instruction-following
quality.

\subsubsection{RL algorithm}
\label{sec:rl-algorithm}

Both stages train the policy with a variant of
GRPO~\cite{shao2024deepseekmath} adapted for asynchronous rollouts and
equipped with stability mechanisms that handle the
train$\leftrightarrow$inference mismatch we see on BF16 + MoE policies.

\paragraph{GRPO loss.}
We use the GRPO recipe with the modifications that have become standard
across recent open RL systems. The loss is token-level: every valid
generated token contributes equally to the gradient, as recommended by
DAPO~\cite{yu2025dapo} and Dr.~GRPO~\cite{liu2025understanding}.
Advantages are computed per prompt group with a leave-one-out baseline
and \emph{without} standard-deviation normalization, again following
Dr.~GRPO. We sample $G$ responses per prompt, oversample by roughly
$1.5\times$, and discard prompt groups whose within-group reward
variance is zero, an approximate version of the dynamic-sampling step
from DAPO. The PPO surrogate uses an asymmetric clip range
$[1 - \epsilon_\text{low},\, 1 + \epsilon_\text{high}]$, the
``clip-higher'' setting introduced by DAPO, which lets positive-advantage
updates flow more freely than negative ones. We do not anchor the policy
to the SFT reference with a KL term; recent large-scale open RL systems
have converged on omitting this
term~\cite{olmo3,yang2025qwen3,nemotronnanov2,yu2025dapo}.

\paragraph{Asynchronous rollouts.}
Rollouts and gradient updates run on different GPUs
(\cref{sec:rl-infra}); the trainer pulls a batch from a
continuously-filling trajectory buffer rather than waiting for
generation. Trajectory staleness is bounded so that a rollout's tokens
are at most two training steps older than the policy used in the
gradient update.

\paragraph{Train versus inference importance sampling.}
Even when the inference policy and the trainer's recomputed policy are
nominally the same model, the two forward passes can disagree on
per-token log-probabilities. The principal source of this non-determinism in
an MoE policy is the router itself: for the same hidden state, the inference-time
router may dispatch a token to a different expert than the trainer-side
router, and the resulting logits and log-probabilities
differ even though the weights are identical. BF16 numerical stability contributes
additional noise. We track this disparity through the train-versus-inference
ratio:
\begin{equation*}
  \rho_t \;=\; \frac{\pi_\text{train}(y_t \mid y_{<t};\, \theta_\text{old})}
                    {\pi_\text{infer}(y_t \mid y_{<t};\, \theta_\text{old})},
\end{equation*}
which is not exactly $1$ even before any gradient update. Left
unbounded in the loss, $\rho_t$ would let a small number of drifted
tokens dominate the gradient. This is distinct from the standard PPO
ratio between the current and pre-step training policies introduced
below; PPO clipping handles the latter, IcePop handles~$\rho_t$.

We use per-token IcePop truncation~\cite{ring1t} to guard against this.
For each generated token we keep its contribution to the loss only when
$\rho_t \in [\alpha, \beta]$; the contribution is set to zero outside
the band. Unlike the PPO clip, which caps an out-of-band ratio at the
clip edge, IcePop drops the token entirely. This is the safer default
when the cause of a large $\rho_t$ is an expert flip rather than a real
on-policy update we want to apply.

Putting the pieces together, the per-step loss minimised by the trainer
is
\begin{equation*}
\begin{aligned}
A_i \;&=\; R_i \;-\; \frac{1}{G-1}\sum_{j \neq i} R_j, \\[2pt]
r_{i,t} \;&=\; \frac{\pi_\text{train}(y_{i,t} \mid y_{i,<t};\, \theta)}
                    {\pi_\text{train}(y_{i,t} \mid y_{i,<t};\, \theta_\text{old})},
\qquad
\rho_{i,t} \;=\; \frac{\pi_\text{train}(y_{i,t} \mid y_{i,<t};\, \theta_\text{old})}
                       {\pi_\text{infer}(y_{i,t} \mid y_{i,<t};\, \theta_\text{old})}, \\[4pt]
M(\rho) \;&=\;
\begin{cases}
\rho & \text{if } \alpha \le \rho \le \beta, \\
0 & \text{otherwise,}
\end{cases} \\[4pt]
\mathcal{L}_\text{GRPO}
\;&=\;
-\,\frac{1}{N_\text{tok}}\,
\sum_{i,t}
M(\rho_{i,t})\,
\min\!\Big(
r_{i,t}\, A_i,\;
\mathrm{clip}\!\big(r_{i,t},\, 1 - \epsilon_\text{low},\, 1 + \epsilon_\text{high}\big)\, A_i
\Big),
\end{aligned}
\end{equation*}
where $r_{i,t}$ is the standard PPO ratio between the trainer's current
and pre-step policies, $\rho_{i,t}$ is the train-versus-inference
disparity that IcePop calibrates, $G$ is the number of responses per
prompt, and $N_\text{tok}$ is the total number of valid generated tokens
in the batch. The four choices that distinguish this recipe from
textbook GRPO are visible in the formula:
\begin{enumerate}
  \item a leave-one-out baseline without standard-deviation
    normalization;
  \item the IcePop calibration $M(\rho_{i,t})$ that zeroes the
    contribution of any token whose train-versus-inference ratio falls
    outside $[\alpha, \beta]$;
  \item token-level normalization by the total valid-token count;
  \item the asymmetric clip-higher range
    $\epsilon_\text{low} < \epsilon_\text{high}$.
\end{enumerate}

\paragraph{Reward shaping.}
We add two reward-shaping rules on top of the verifier's raw score.

The first is the soft overlong penalty from DAPO~\cite{yu2025dapo}.
Rewards inside a buffer region just below the maximum response length
interpolate linearly between the raw score at the buffer's lower edge
and a configured floor at the length cap; rollouts that exceed the cap
are dropped from the loss entirely, also following DAPO. This avoids
training on samples that simply ran out of budget while preserving the
gradient signal on shorter samples.

The second is a concision penalty applied selectively to non-thinking
responses. During an early Instruct run we observed that the policy
began producing inline reasoning without the \texttt{<think>}
delimiters used by the Thinking variant, contradicting the deployment
contract of a brief Instruct model. Late-training math rollouts
looked like the following:

\begin{rolloutquote}
[\ldots] \textbf{But wait}, I recall that in some similar problems,
the answer is more than $3$. \textbf{Wait}, let me check online or
think again. \textbf{Wait}, perhaps I missed a case. \textbf{Wait},
what if the number is of the form $p^4 q^2$, but with the same prime?
No, then it would be $p^6$, which has $7$ divisors, not $15$. So no.
\textbf{Wait}, but let's check $n = 144$, $400$, $324$, all less than
$500$. [\ldots]
\end{rolloutquote}

Models tend to mark such reasoning with a fairly stable lexicon of
trigger words (\emph{wait}, \emph{actually}, \emph{hmm}, \emph{let me
think}, and similar markers); we follow the ARLCP-style penalty
of~\cite{arlcp2026} and multiplicatively shrink the reward on correct
rollouts in proportion to the number of trigger words present in the
response. The multiplier is bucketed into three tiers of increasing
strength as the trigger count grows, and we apply the penalty only on
tasks where the lexicon is not legitimately part of the output, so
that thinking-mode responses on math and reasoning tasks are not
penalised. The penalty drives the leakage down sharply at the
population level: in math rollouts sampled near the end of training,
the average rollout in the no-concision run carried $7.3$
reflection-trigger words ($0.75$ per $1000$ characters of response),
against $0.6$ ($0.21$ per $1000$ characters) in the production
Instruct run with the penalty enabled.

\subsubsection{Training Setup}
\label{sec:rl-setup}

Both stages share the optimizer recipe and overall training loop. The
trainer uses distributed AdamW with peak learning rate $1\!\times\!10^{-6}$,
decaying to $1\!\times\!10^{-7}$, with a linear warmup over the first
50 iterations and a constant schedule for the remainder of the run. We
keep the BF16 / FP8 hybrid precision recipe from pre-training
(\cref{sec:precision}), and clip gradients at norm $1.0$.
\Cref{tab:rl-hparams} lists the per-stage hyperparameters; the
dominant differences between the two runs are the sequence budget and
the number of training steps.

\begin{table}[H]
  \centering
  \caption{Per-stage RL hyperparameters. Shared rows apply to both
  runs; rows below the rule differ between Instruct and Thinking.}
  \label{tab:rl-hparams}
  \small
  \begin{tabular}{@{}lcc@{}}
    \toprule
    \tableheadercolor
    \textbf{Hyperparameter} & \textbf{Instruct} & \textbf{Thinking} \\
    \midrule
    Prompts per step                                          & \multicolumn{2}{c}{256} \\
    Generations per prompt                                    & \multicolumn{2}{c}{16} \\
    Global batch size                                         & \multicolumn{2}{c}{4{,}096} \\
    Oversampling factor                                       & \multicolumn{2}{c}{$1.5\times$} \\
    Trajectory age cap                                        & \multicolumn{2}{c}{2 steps} \\
    PPO clip $\epsilon_\text{low}$ / $\epsilon_\text{high}$    & \multicolumn{2}{c}{$0.2$ / $0.28$} \\
    IcePop band $[\alpha, \beta]$                             & \multicolumn{2}{c}{$[0.5, 5.0]$} \\
    KL coefficient                                            & \multicolumn{2}{c}{$0$} \\
    Optimizer                                                 & \multicolumn{2}{c}{AdamW, $(\beta_1, \beta_2) = (0.9, 0.999)$, $\text{wd} = 0.01$} \\
    Peak / min learning rate                                  & \multicolumn{2}{c}{$1\!\times\!10^{-6}$ / $1\!\times\!10^{-7}$} \\
    LR schedule                                               & \multicolumn{2}{c}{constant, 50-iter linear warmup} \\
    Max gradient norm                                         & \multicolumn{2}{c}{$1.0$} \\
    Max rollout turns                                         & \multicolumn{2}{c}{10} \\
    \midrule
    Trainer micro-batch size                                  & 2                & 1 \\
    Max total sequence length (tokens)                        & 16{,}384         & 40{,}960 \\
    Training steps                                            & 500              & 100 \\
    \bottomrule
  \end{tabular}
\end{table}

\begin{figure}[H]
\centering
\includegraphics[width=0.85\textwidth]{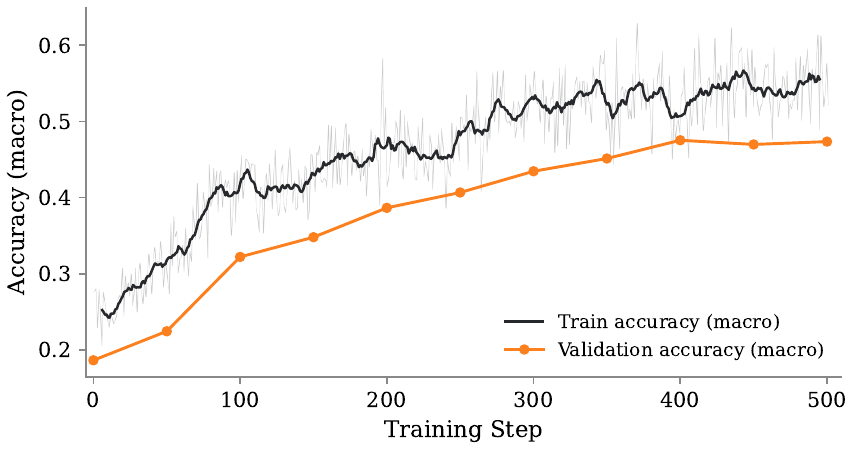}
\caption{Training and validation accuracy (macro-averaged across
tasks) for the Instruct RL run. The smoothed train curve is shown in
black with the raw per-step values rasterised underneath; validation
is sampled every 50 steps.}
\label{fig:rl-instruct-accuracy}
\end{figure}

\paragraph{Instruct.}
The Instruct stage starts from the SFT-Instruct
checkpoint (\cref{sec:sft}) and trains on the Instruct data mix
(\cref{tab:rl-mix}) for 500 steps. The shorter response budget
allows two rollouts per trainer micro-batch and a maximum total
sequence length of 16{,}384 tokens.
\Cref{fig:rl-instruct-accuracy} shows the train and validation
accuracy curves for this run.

\paragraph{Thinking.}
The Thinking stage is a cold restart from the SFT-Thinking checkpoint
(\cref{sec:sft}) and trains on the Thinking data mix
(\cref{tab:rl-mix}) for 100 steps. To accommodate long chains of
thought we lift the maximum total sequence length to 40{,}960 tokens,
which forces the trainer's micro-batch size down to one.

\subsection{Post-Training Evaluation}
\label{sec:posttrain-eval}

We evaluate post-trained variants of \modelname against a panel of open-weight
models in the 4B--14B range: Qwen3.5-4B and Qwen3.5-9B~\cite{yang2025qwen3},
OLMo-3-7B~\cite{olmo3}, Ministral-3-14B~\cite{liu2026ministral3}, and
Seed-Coder-8B~\cite{zhang2025seedcoder}. We report two tables: one comparing
the \emph{instruct} (no-thinking) variants in \Cref{tab:posttrain-eval-instruct},
and one comparing the \emph{thinking}/reasoning variants in
\Cref{tab:posttrain-eval-thinking}. 

The post-training evaluation suite covers seven capability areas:
\begin{itemize}
  \item \textbf{Coding:} 
    LiveCodeBench v6~\cite{jain2024livecodebench} (scored over all v1-6 cohorts), 
    EvalPlus (the average of HumanEval+ and MBPP+)~\cite{liu2023evalplus}, and
    MultiPL-E~\cite{cassano2022multipl} (restricted to 7 of the 18 languages
in the original suite: \texttt{C++}, \texttt{Java}, \texttt{PHP},
\texttt{TypeScript}, \texttt{C\#}, \texttt{Shell}, \texttt{JavaScript}).
  \item \textbf{Tool Use:} BFCL v3 focuses on \textit{multi-turn} function-calling, and v4 extends this with \textit{agentic} capabilities consisting of web-search and memory tools~\cite{patil2025bfcl}.
  \item \textbf{Math:} AIME (average of AIME 2025 and 2026, 30
    questions each) and GSM-Plus~\cite{li2024gsmplus}.
  \item \textbf{Knowledge:} MMLU-Redux~\cite{gema2024mmluredux} and
    GPQA Diamond~\cite{rein2023gpqa}.
  \item \textbf{Conversational:} 
    IFEval~\cite{zhou2023ifeval} (prompt-level strict accuracy),
    MixEval~\cite{ni2024mixeval}, BS-Bench (false premise detection rate), 
    and a JetBrains internal pairwise win rate against Qwen2.5-7B-Instruct.
  \item \textbf{Safety:} HarmBench~\cite{mazeika2024harmbench} (harmful rate, lower is
    better) and XSTest~\cite{rottger2024xstest} (safe compliance rate).
\end{itemize}

LLM-as-a-Judge benchmarks (BS-Bench, JetBrains pairwise, HarmBench, and XSTest) use GPT-5.2 as a judge model. 
All benchmarks run at 0.0 temperature, except for BFCL at 0.01 and LiveCodeBench at 0.2.
All models use greedy decoding.

\begin{table}[H]
\centering
\caption{Post-training evaluation, \textbf{instruct (no-thinking)} variants.
All values are percentages; higher is better except HarmBench (lower is
better). EvalPlus is the average of HumanEval+ and MBPP+. AIME is the
average of AIME 2025 and AIME 2026 (30 questions each). BFCL v4 is the
macro-average of its five subtasks (v1, v2, v3, web search, memory).
JetBrains internal scores are pairwise win rates against
Qwen2.5-7B-Instruct. Em-dashes (---) indicate lacking native tool calling for Seed-Coder-8B.}
\label{tab:posttrain-eval-instruct}
\scriptsize
\setlength{\tabcolsep}{4pt}
\begin{tabular}{@{}l >{\columncolor{mellum-light}}c >{\columncolor{mellum-light}}c@{\hspace{12pt}} ccccc@{}}
\toprule
 & \multicolumn{2}{c}{\textsc{Mellum 2}} & & & & & \\[-2pt]
\cmidrule(lr){2-3}
\textsc{Benchmark} & \textsc{SFT} & \textsc{RL} & \textsc{Qwen3.5-4B} & \textsc{Qwen3.5-9B} & \textsc{OLMo-3-7B} & \textsc{Ministral-3-14B} & \textsc{Seed-Coder-8B} \\
 & {\scriptsize\itshape\color{mellum-gray}2.5B/12B} & {\scriptsize\itshape\color{mellum-gray}2.5B/12B} & {\scriptsize\itshape\color{mellum-gray}4B} & {\scriptsize\itshape\color{mellum-gray}9B} & {\scriptsize\itshape\color{mellum-gray}7B} & {\scriptsize\itshape\color{mellum-gray}14B} & {\scriptsize\itshape\color{mellum-gray}8B} \\
\midrule
\multicolumn{8}{@{}l}{\textit{\color{mellum-gray}Coding}} \\
\addlinespace[2pt]
LiveCodeBench v6      & 30.9 & 37.2 & 51.0 & 63.7 & 28.2 & 42.4 & 28.1 \\
EvalPlus              & 76.2 & 78.4 & 69.4 & 71.8 & 67.3 & 74.1 & 73.8 \\
MultiPL-E$^{\dagger}$ & 64.6 & 67.1 & 51.0 & 67.1 & 36.1 & 71.5 & 77.0 \\
\addlinespace[4pt]
\multicolumn{8}{@{}l}{\textit{\color{mellum-gray}Tool Use}} \\
\addlinespace[2pt]
BFCL v4               & 31.8 & 44.2 & 52.0 & 60.6 & 19.8 & 38.8 & --- \\
BFCL v3               & 43.1 & 66.3 & 64.1 & 70.5 & 41.9 & 52.7 & --- \\
\addlinespace[4pt]
\multicolumn{8}{@{}l}{\textit{\color{mellum-gray}Math}} \\
\addlinespace[2pt]
AIME                  & 29.9 & 41.7 & 38.3 & 58.3 & 40.0 & 33.3 &  0.0 \\
GSM-Plus              & 73.0 & 80.5 & 85.2 & 87.9 & 85.8 & 86.6 & 50.4 \\
\addlinespace[4pt]
\multicolumn{8}{@{}l}{\textit{\color{mellum-gray}Knowledge}} \\
\addlinespace[2pt]
MMLU-Redux            & 77.4 & 78.1 & 87.5 & 91.1 & 71.8 & 85.9 & 38.1 \\
GPQA Diamond          & 38.9 & 40.9 & 76.8 & 79.8 & 40.9 & 58.6 & 20.2 \\
\addlinespace[4pt]
\multicolumn{8}{@{}l}{\textit{\color{mellum-gray}Conversational}} \\
\addlinespace[2pt]
IFEval                & 69.3 & 75.8 & 82.1 & 83.9 & 83.2 & 67.3 & 56.2 \\
JetBrains pairwise    & 66.7 & 68.1 & 60.6 & 77.8 & 44.4 & 72.4 & 43.0 \\
MixEval               & 62.9 & 62.2 & 65.9 & 71.1 & 59.4 & 71.2 & 37.2 \\
BS-Bench              & 24.0 & 18.0 & 56.9 & 61.0 & 22.0 &  9.0 &  5.0 \\
\addlinespace[4pt]
\multicolumn{8}{@{}l}{\textit{\color{mellum-gray}Safety}} \\
\addlinespace[2pt]
HarmBench ($\downarrow$) &  8.4 & 23.1 & 20.3 & 20.9 & 14.7 & 56.5 & 40.0 \\
XSTest                   & 78.3 & 81.2 & 93.2 & 91.2 & 91.2 & 96.8 & 86.3 \\
\bottomrule
\end{tabular}
\end{table}

\begin{table}[H]
\centering
\caption{Post-training evaluation, \textbf{thinking}/reasoning variants.
Same metric conventions as \Cref{tab:posttrain-eval-instruct}. OLMo-3-7B-Thinking does not support native tool calling.}
\label{tab:posttrain-eval-thinking}
\footnotesize
\setlength{\tabcolsep}{4pt}
\begin{tabular}{@{}l >{\columncolor{mellum-light}}c >{\columncolor{mellum-light}}c@{\hspace{12pt}} cccc@{}}
\toprule
 & \multicolumn{2}{c}{\textsc{Mellum 2}} & & & & \\[-2pt]
\cmidrule(lr){2-3}
\textsc{Benchmark} & \textsc{SFT} & \textsc{RL} & \textsc{Qwen3.5-4B} & \textsc{Qwen3.5-9B} & \textsc{OLMo-3-7B} & \textsc{Ministral-3-14B} \\
 & {\scriptsize\itshape\color{mellum-gray}2.5B/12B} & {\scriptsize\itshape\color{mellum-gray}2.5B/12B} & {\scriptsize\itshape\color{mellum-gray}4B} & {\scriptsize\itshape\color{mellum-gray}9B} & {\scriptsize\itshape\color{mellum-gray}7B} & {\scriptsize\itshape\color{mellum-gray}14B} \\
\midrule
\multicolumn{7}{@{}l}{\textit{\color{mellum-gray}Coding}} \\
\addlinespace[2pt]
LiveCodeBench v6      & 75.1 & 69.9 & 59.4 & 68.3 & 59.8 & 42.7 \\
\addlinespace[4pt]
\multicolumn{7}{@{}l}{\textit{\color{mellum-gray}Tool Use}} \\
\addlinespace[2pt]
BFCL v4               & 38.8 & 45.6 & 42.9 & 42.7 & --- & 35.9 \\
BFCL v3               & 60.5 & 69.4 & 73.9 & 68.5 & --- & 52.2 \\
\addlinespace[4pt]
\multicolumn{7}{@{}l}{\textit{\color{mellum-gray}Math}} \\
\addlinespace[2pt]
AIME                  & 20.0 & 58.4 & 68.3 & 73.4 & 61.7 & 38.3 \\
GSM-Plus              & 62.6 & 87.0 & 89.3 & 90.7 & 88.1 & 86.5 \\
\addlinespace[4pt]
\multicolumn{7}{@{}l}{\textit{\color{mellum-gray}Knowledge}} \\
\addlinespace[2pt]
MMLU-Redux            & 84.8 & 86.2 & 88.3 & 91.7 & 71.3 & 84.4 \\
GPQA Diamond          & 39.9 & 57.6 & 76.8 & 81.3 & 29.3 & 46.0 \\
\addlinespace[4pt]
\multicolumn{7}{@{}l}{\textit{\color{mellum-gray}Conversational}} \\
\addlinespace[2pt]
IFEval                & 69.1 & 76.5 & 87.1 & 89.8 & 84.7 & 59.7 \\
JetBrains pairwise    & 64.4 & 69.5 & 40.5 & 56.7 & 32.2 & 63.8 \\
MixEval               & 63.4 & 66.9 & 71.9 & 76.0 & 67.0 & 70.8 \\
BS-Bench              & 14.0 & 15.0 & 63.0 & 70.0 & 23.0 &  9.0 \\
\addlinespace[4pt]
\multicolumn{7}{@{}l}{\textit{\color{mellum-gray}Safety}} \\
\addlinespace[2pt]
HarmBench ($\downarrow$) & 12.2 & 20.6 & 15.9 &  6.6 &  48.7 & 70.0 \\
XSTest                   & 90.8 & 89.6 & 96.8 & 97.6 &  93.2 & 96.8 \\
\bottomrule
\end{tabular}
\end{table}

\paragraph{Overall profile.}
The seven capability areas reveal a consistent picture: \modelname is
strongest where the domain aligns with our training mix (function-level
code synthesis and JetBrains-style developer interaction),
competitive on tool use and math once RL is applied, and weakest on
broad world knowledge. With only 2.5B active parameters drawn from a
12B MoE backbone, the model is competing against dense baselines that
range from 4B (Qwen3.5-4B) to 14B (Ministral-3-14B); we contextualize
the results in that light below.

\paragraph{Coding.}
The three coding benchmarks measure different abilities and the
results separate cleanly. \textbf{EvalPlus} -- the augmented
HumanEval+/MBPP+ pair that probes robust function-level synthesis --
is led by \modelname-RL at 78.4\%, ahead of every baseline including
Qwen3.5-9B (71.8) and the code-specialized Seed-Coder-8B (73.8). This
is the regime our pre-training mix targets directly.
\textbf{LiveCodeBench v6}, by contrast, draws on
contamination-resistant competitive-programming problems that demand
multi-step algorithmic reasoning over relatively few tokens; the
instruct variant lags the Qwen3.5 series (37.2 vs.\ 51.0 / 63.7) but
matches or beats the other 7--14B baselines. The gap closes
dramatically in the thinking configuration: \modelname-SFT-Thinking
reaches 75.1, the top score in our panel and 6.8 points ahead of
Qwen3.5-9B-Thinking. We read this as evidence that algorithmic
reasoning is in the model's reach but requires an explicit thinking
budget to be unlocked, whereas function synthesis transfers from
pre-training without one. \textbf{MultiPL-E}, restricted here to seven
of the eighteen native languages, is mid-pack: Seed-Coder-8B (77.0)
and Ministral-3-14B (71.5) edge ahead on cross-lingual breadth.

\paragraph{Tool use, math, and reasoning.}
RL is where the largest single-step jumps appear. BFCL v3 climbs from
43.1 to 66.3 (instruct) and 60.5 to 69.4 (thinking), with the thinking
variant overtaking Qwen3.5-9B-Thinking (68.5). On BFCL v4, which adds
agentic web-search and memory subtasks, \modelname-RL-Thinking leads
the panel at 45.6, against 42.9 / 42.7 for the Qwen3.5 family --- a
sign that our function-calling RL recipe transfers usefully to
held-out agentic settings. Math follows a similar arc: AIME goes from
29.9 (SFT instruct) to 41.7 (RL instruct) and from 20.0 to 58.4 in
thinking mode. The SFT-Thinking AIME score is below its SFT-instruct
counterpart, a quirk we attribute to the thinking head requiring
RL-stage exposure to mathematical reasoning before its reasoning
trace is well-calibrated for that task family. GSM-Plus reaches 87.0
in RL-Thinking, within a few points of Qwen3.5-9B-Thinking (90.7).

\paragraph{Knowledge: the principal weakness.}
MMLU-Redux and GPQA Diamond are the area where the Qwen3.5 series is
dominant: 91.1 / 79.8 at 9B against our 78.1 / 40.9 (instruct) and
86.2 / 57.6 (thinking). GPQA in particular --- graduate-level science
QA --- is essentially a probe of factual depth outside computer
science, and the gap reflects a deliberate tradeoff in our training
mix toward code and developer documentation rather than broad
encyclopedic coverage. For a code-assistant model this profile is
acceptable, but it bounds the off-domain use of \modelname and is
worth surfacing explicitly to deployers.

\paragraph{Conversational: JetBrains-relative leadership, generic mid-pack.}
On the internal JetBrains pairwise win-rate against
Qwen2.5-7B-Instruct, \modelname-RL-Thinking leads the panel at 69.5\%,
above both Ministral-3-14B-Thinking (63.8) and Qwen3.5-9B-Thinking
(56.7), while on the generic conversational benchmarks (IFEval,
MixEval) the model sits in the middle of the pack. The asymmetry is
informative: the pairwise judge sees code-aware, developer-flavored
prompts where domain familiarity pays off, whereas the generic
benchmarks reward broad-coverage chat behavior that benefits from the
Qwen3.5 post-training mix. BS-Bench is the conversational outlier:
\modelname scores 14--24 against 56--70 for the Qwen3.5 series. This
benchmark rewards push-back against false premises rather than helpful
task completion; the gap suggests our SFT/RL signal leans toward
compliance, and we leave tightening this trade-off for future
iterations.

\paragraph{Safety.}
On HarmBench (lower is better), \modelname-SFT is the safest model in
the instruct table at 8.4\%, with Ministral-3-14B (56.5) and
Seed-Coder-8B (40.0) substantially worse. The RL variant regresses to
23.1\%, consistent with the well-documented tendency of
preference-optimization stages to relax some refusal behaviors;
this is a known alignment tax in our RL recipe and a target for
future iterations. On
XSTest, \modelname trails the largest baselines by roughly ten points,
indicating that a subset of safe prompts are over-refused; we view
this as the symmetric counterpart to the HarmBench regression and an
item for joint optimization in subsequent releases.

\section{Efficiency and Deployment}
\label{sec:efficiency}

Practical deployment in latency-sensitive IDE environments is a core design goal
of \modelname. The architecture was designed from the outset to match or exceed
the inference speed of Qwen2.5-7B~\cite{qwen2.5}.

We built a dedicated inference benchmarking pipeline with fixed hardware,
software dependencies, and Docker containers to ensure reproducibility across
all architectural candidates. Benchmarks use representative input/output sizes
from production code completion workloads (mean input length of 2{,}304 tokens,
mean output length of 256 tokens) and evaluate in two regimes: \emph{sync
mode}, which measures sequential single-request latency, and
\emph{throughput mode}, which measures sustained tokens/s under concurrent
high-load requests. Throughput mode uses no fixed request rate: the client
issues requests back-to-back to keep the server saturated, and the sustained
rates we measure are 20.2~req/s for \modelname, 16.7~req/s for Qwen2.5-7B,
and 11.3~req/s for Qwen3-8B. All measurements use a single H100 GPU (80\,GB) with
vLLM~\cite{kwon2023vllm} serving and dynamic FP8 model quantization on a host with 192\,GB
of system RAM and 48 CPU cores.

\begin{figure}[H]
  \centering
  \includegraphics[width=0.95\textwidth]{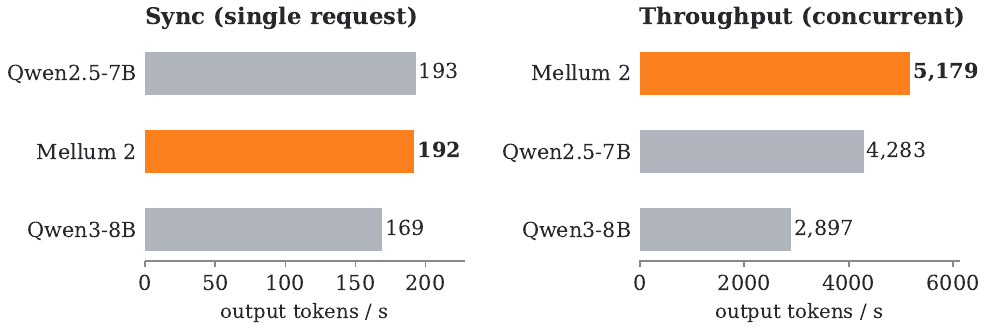}
  \caption{Output tokens/s on a single H100, vLLM FP8 serving, at the
    benchmark workload shape (ISL/OSL\,=\,2{,}304/256). \modelname matches the
    sync latency of Qwen2.5-7B while delivering 21\% higher sustained
    throughput.}
  \label{fig:throughput-comparison}
\end{figure}

\Cref{fig:throughput-comparison} compares \modelname against two dense
baselines, Qwen2.5-7B~\cite{qwen2.5} and Qwen3-8B~\cite{yang2025qwen3}. In sync
mode, \modelname matches the 193~tokens/s of Qwen2.5-7B---the architectural
target set in \Cref{sec:arch-decisions}---to within a single token. In
throughput mode, it pulls 21\% ahead of Qwen2.5-7B and 79\% ahead of Qwen3-8B.

\section{Conclusion}
\label{sec:conclusion}

We have presented \modelname, an open-weight 12B-parameter Mixture-of-Experts
model with 2.5B active parameters, released as matched \emph{Instruct} and
\emph{Thinking} variants under the Apache~2.0 license. As the general-purpose
successor to the 4B dense \prevmodelname completion model, it is built to
generate and edit code, reason through engineering tasks, call tools, and
drive agentic workflows inside the IDE at a per-token cost that is practical
to deploy at scale.

Every architectural decision, including MoE versus dense, 8-of-64 expert sparsity, 4-KV-head GQA, the 3:1 Sliding Window Attention pattern, and the single MTP head, was selected by ablation under a fixed inference budget: matching the single-H100 speed of Qwen2.5-7B.
The resulting model meets that target in
single-request decoding (192 vs.\ 193 tokens/s) and exceeds it by 21\,\%
under concurrent serving (5{,}179 tokens/s). On top of this, we ran a
three-phase pre-training curriculum on ${\sim}$10.65T tokens with a
Muon\,+\,FP8-hybrid stack, extended context to 131{,}072 tokens via
layer-selective YaRN, and applied a two-stage post-training pipeline (SFT
followed by RLVR on math and executable coding). Across code, math, tool
use, knowledge, conversational, and safety benchmarks, \modelname is
competitive with open-weight baselines in the 4--14B range while running at
the per-token compute of a 2.5B dense model.

Natural directions to explore from here include:
\begin{enumerate}
  \item pushing \modelname further into SWE RL---training directly on
    repository-level software-engineering tasks and toward competitive
    small SWE agents;
  \item broader scaling of RL infrastructure and environment coverage;
  \item revisiting the long-context mid-training mix.
\end{enumerate}
Looking further out, the same recipe of selecting architecture by
ablation against a fixed inference budget also opens the door to a
larger, similarly inference-aware Mellum.

We release the base, instruct, and thinking checkpoints together with
this report,
with the aim of giving the community both an open recipe and an
inference-aware design point for small-MoE coding models.

\newpage
\printbibliography[title={References}]

\appendix

\section{Architecture Exploration Details}
\label{app:arch-exploration}

This appendix provides additional detail on the architecture exploration
experiments summarized in \cref{sec:arch-decisions}.

\subsection{Dense Architecture Exploration}

We evaluated dense architectures based on Qwen3~\cite{yang2025qwen3} variations across two axes:

\textbf{Deeper variants} (32--40 layers, hidden size 3072--4096): None
consistently outperformed Qwen2.5-7B~\cite{qwen2.5} on evaluation benchmarks under the
latency constraint. Deeper architectures suffer from more sequential
operations, degrading inference performance.

\textbf{Wider variants} (24--28 layers, hidden size 3584--4096): Wider and
shallower architectures exhibited better inference performance, as expected,
but still failed to consistently exceed the Qwen2.5-7B quality baseline.

\textbf{Multi-head Latent Attention (MLA)}~\cite{deepseekai2024deepseekv2}: We adapted the DeepSeek
architecture by removing MoE layers and enabling MLA. With a latent rank of
512 (the only rank supported by the vLLM~\cite{kwon2023vllm} inference backend at the time), MLA
allowed scaling to approximately 5.5B parameters at Qwen2.5-7B latency.
However, quality improvements were insufficient, and the latent rank was
overly large for our model scale, limiting the potential KV-cache savings.

\subsection{MoE Architecture Exploration}

We scaled down the Qwen3-30B-A3B~\cite{yang2025qwen3} architecture proportionally while preserving
the ratios between hidden size, intermediate size, and expert size. Key
findings:

\begin{itemize}
  \item \textbf{Expert count}: Fixed at 64 (maximum that fits in GPU memory).
  \item \textbf{Active experts}: 2 active experts achieved ${\sim}$1.5$\times$
    lower latency than 8, but quality was substantially worse at our model
    scale. 8 active experts provided the best quality--latency trade-off.
  \item \textbf{Total parameters}: Up to ${\sim}$15B total parameters were
    feasible while matching Qwen2.5-7B latency with 8 active experts.
  \item \textbf{Shared expert}: Adding a shared expert~\cite{dai2024deepseekmoe} (always active in
    addition to the routed top-$k$) yielded no measurable quality gain at our
    scale and consistently hurt inference performance because of the extra
    always-on FFN compute per token. We dropped it from the final design.
  \item \textbf{Dense/sparse interleaving}: Replacing a subset of MoE layers
    with dense FFN layers (in the spirit of recent
    interleaved-MoE designs) similarly hurt inference performance without a
    matching quality improvement, so all FFN layers in \modelname{} are MoE.
  \item \textbf{Auxiliary-loss-free load balancing}: We were strongly tempted
    to adopt the auxiliary-loss-free, bias-based load-balancing scheme
    popularised by DeepSeek-V3~\cite{deepseekai2025deepseekv3}: it simplifies
    the training stack by removing an extra loss term and its coefficient,
    and in our short-run experiments it matched or slightly improved expert
    utilisation. We ultimately stayed with the auxiliary-loss formulation in
    order to fit cleanly into the Qwen3-MoE module layout, which is what
    every major open-source inference framework already implements; this made
    integration of \modelname{} into the existing ecosystem essentially
    free. We plan to switch to auxiliary-loss-free balancing in the next
    iteration, once the loss-free variant is equally well supported
    downstream.
\end{itemize}

\subsection{Hybrid Architecture Exploration}
\label{app:hybrid-arch}

In parallel with the dense and MoE sweeps above, we also explored
\emph{hybrid} attention designs that interleave standard softmax attention
with linear-recurrent token mixers. Concretely, we built variants based on
the Qwen3-Next recipe~\cite{qwen2026qwen35} (later adopted in the Qwen3.5
family), which replaces a large fraction of attention layers with Gated
DeltaNet~\cite{yang2025gateddeltanet} layers, keeping only every fourth
layer as full attention.

On long-context, large-batch workloads are very attractive for such hybrids: the
fixed-size recurrent state of Gated DeltaNet eliminates the linearly growing
KV cache and gives near-constant per-token decode cost. For \modelname{},
however, the dominant deployment target is \emph{short context, single batch}
in-IDE inference, where the scenario is inverted. At the time we ran our
architecture search, every hybrid variant we benchmarked exhibited a substantial latency regression on short
input/output lengths compared with a pure-attention baseline of the same
parameter budget. The reasons are at least partly structural: the recurrent
state update is more arithmetically heavy than a standard attention step at
small sequence lengths, decode is memory-bound on the state matrix rather
than on a tiny KV cache, and the relevant kernels were significantly less
optimised in mainstream inference backends than the long-standing softmax
attention path.

Because none of these issues are fundamental --- they reflect kernel and
framework maturity rather than the underlying algorithm --- we expect the
short-context inference gap to shrink as hybrid architectures see wider
adoption and dedicated optimisation in inference engines, and we intend to
revisit hybrid designs for future \modelname{} iterations.

\subsection{MoE Training Hyperparameters}
\label{app:moe-training}

We conducted preliminary experiments on MoE-specific hyperparameters before
the main training sweeps:

\begin{itemize}
  \item \textbf{Balancing strategy}: Per-sequence auxiliary loss produced
    slightly better test loss than global-batch balancing on short runs.
    We selected global-batch balancing for its flexibility with variable
    batch sizes.
  \item \textbf{Auxiliary loss coefficient}: $10^{-2}$ performed better on
    short runs, but we chose $10^{-3}$ for full pre-training to avoid
    over-constraining expert utilization.
  \item \textbf{Token dropping}: Experiments with expert capacity factors of
    1.0--1.5 showed no meaningful quality difference. We adopted dropless
    routing, which was initially slower but improved in throughput as the
    router learned to balance load during training. The residual overhead
    is ${\sim}$15\% at the time of writing.
\end{itemize}

\section{Training Hyperparameters (Full)}
\label{app:hyperparams}

\begin{table}[H]
  \centering
  \caption{Complete training hyperparameters for \modelname pre-training.}
  \label{tab:full-hyperparams}
  \small
  \setlength{\tabcolsep}{4pt}
  \begin{minipage}[t]{0.48\textwidth}
    \centering
    \begin{tabular}{@{}ll@{}}
      \toprule
      \tableheadercolor
      \multicolumn{2}{@{}l}{\textbf{\textsc{Optimizer}}} \\
      \midrule
      Optimizer                      & Distributed Muon \\
      Muon momentum                  & 0.95 \\
      Muon Newton--Schulz iterations & 5 \\
      Muon scale mode                & spectral \\
      Muon TP mode                   & blockwise \\
      Muon extra scale factor        & 0.2 \\
      Nesterov momentum              & Yes \\
      Adam $\beta_1, \beta_2$        & 0.9, 0.95 \\
      Adam $\epsilon$                & $10^{-8}$ \\
      SGD momentum                   & 0.9 \\
      Weight decay                   & 0.1 \\
      Gradient clipping              & 1.0 \\
      \addlinespace[4pt]
      \toprule
      \tableheadercolor
      \multicolumn{2}{@{}l}{\textbf{\textsc{Learning Rate}}} \\
      \midrule
      Peak learning rate             & $3 \times 10^{-4}$ \\
      Minimum learning rate          & 0 \\
      Schedule                       & WHD \\
      Warmup steps                   & 2{,}000 (linear) \\
      Decay steps                    & 49{,}306 (linear) \\
      Decay style                    & Linear \\
      \addlinespace[4pt]
      \toprule
      \tableheadercolor
      \multicolumn{2}{@{}l}{\textbf{\textsc{Batch \& Sequence}}} \\
      \midrule
      Sequence length                & 8{,}192 \\
      Global batch size              & 4{,}096 sequences \\
      Micro batch size               & 2 \\
      Batch rampup                   & 2{,}048 $\to$ 4{,}096 \\
      Total training steps           & 323{,}459 \\
      \bottomrule
    \end{tabular}
  \end{minipage}\hfill
  \begin{minipage}[t]{0.48\textwidth}
    \centering
    \begin{tabular}{@{}ll@{}}
      \toprule
      \tableheadercolor
      \multicolumn{2}{@{}l}{\textbf{\textsc{Precision}}} \\
      \midrule
      Base precision                 & BF16 \\
      FP8 mode                       & Hybrid \\
      FP8 recipe                     & Tensorwise \\
      FP8 amax algorithm             & Most recent \\
      FP8 parameter gather           & Yes \\
      Gradient reduction             & FP32 \\
      \addlinespace[4pt]
      \toprule
      \tableheadercolor
      \multicolumn{2}{@{}l}{\textbf{\textsc{MoE}}} \\
      \midrule
      Auxiliary loss type            & Global batch \\
      Auxiliary loss coefficient     & $10^{-3}$ \\
      Z-loss coefficient             & $10^{-3}$ \\
      Router bias update rate        & $10^{-3}$ \\
      Router precision               & FP32 \\
      Token dropping                 & Disabled \\
      Grouped GEMM                   & Yes \\
      Router fusion                  & Yes \\
      Permute fusion                 & Yes \\
      \addlinespace[4pt]
      \toprule
      \tableheadercolor
      \multicolumn{2}{@{}l}{\textbf{\textsc{Parallelism}}} \\
      \midrule
      Expert parallelism             & 8 \\
      Tensor parallelism             & 1 \\
      Pipeline parallelism           & 1 \\
      \addlinespace[4pt]
      \toprule
      \tableheadercolor
      \multicolumn{2}{@{}l}{\textbf{\textsc{Multi-Token Prediction}}} \\
      \midrule
      Additional prediction layers   & 1 \\
      MTP loss scaling factor        & 0.1 \\
      \bottomrule
    \end{tabular}
  \end{minipage}
\end{table}

\section{Evaluation Notes and Lessons Learned}
\label{app:eval-notes}

This appendix collects two evaluation-time observations that shaped how we
report numbers in \cref{sec:posttrain-eval} and that we believe are useful for other
groups running similar pipelines.

\subsection{RULER QA Subsets and Prompt Formatting}
\label{app:ruler-formatting}

Throughout the long-context extension stage (\cref{sec:long-context}), we
used RULER~\cite{hsieh2024ruler} at 128K as the primary long-context
benchmark. Early in the run, we observed that the model scored approximately
zero on the QA subsets while behaving normally on the retrieval and
aggregation tasks. The failure mode was not a capability gap: the model
was \emph{continuing} the question (generating plausible follow-up
questions in the same style) rather than answering it, and the
exact-match scorer counted every such response as wrong.

The lower quality resulted from a prompt-formatting issue rather than an actual capability gap. We
deliberately did not add RULER-style QA prompts to the long-context data
mix, since doing so would have amounted to optimizing for the benchmark
rather than for the underlying capability.

\subsection{Reasoning Budgets for Qwen3 and Qwen3.5 Thinking Variants}
\label{app:qwen-thinking-budget}

While evaluating the \emph{thinking} variants of Qwen3-4B and Qwen3.5-4B
(reported in \cref{tab:posttrain-eval-thinking}), we encountered a
consistent failure mode on a non-trivial fraction of prompts: the model
would not emit a closing \texttt{</think>} tag and continued to reason
indefinitely. Running these models without a generation cap is both
expensive and produces near-zero benchmark scores, 
because the model would rather fill its context window with a reasoning trace 
than answer the benchmark question. 

Recent vLLM~\cite{kwon2023vllm} releases expose a configurable reasoning
budget that forces the model out of the thinking phase after a chosen
number of tokens. Qwen does not publish an official threshold for the
4B/9B thinking variants, so we used a generous but arbitrary budget of
32K tokens for every thinking model in our evaluation. This is sufficient
to admit long but bounded chains of thought while preventing the
pathological non-terminating cases from dominating the average.

We note that, from a downstream-user perspective, the small thinking
variants of Qwen3 and Qwen3.5 are difficult to deploy in their thinking
regime without such a cap. We do not have a definitive explanation for
this behavior, but we suspect a lack of on-policy reinforcement-learning
training at the smallest scales in those families, since the larger
models in the same families appear to terminate reasoning much more
reliably.

\end{document}